\pdfoutput=1

\documentclass[11pt]{article}

\usepackage[final]{acl}
\usepackage{times}
\usepackage{latexsym}
\usepackage[T1]{fontenc}
\usepackage[utf8]{inputenc}
\usepackage{microtype}
\usepackage{inconsolata}
\usepackage{multicol}
\usepackage{multirow}
\usepackage{booktabs}
\usepackage{amsfonts}
\usepackage{hyperref}
\usepackage{amssymb}
\usepackage{graphicx}
\usepackage{relsize}
\usepackage{tcolorbox}
\usepackage{changepage}
\newtcbox{\mybox}[1][red]
  {on line, arc = 0pt, outer arc = 0pt,
    colback = #1!10!white, colframe = #1!50!black,
    boxsep = 0pt, left = 1pt, right = 1pt, top = 1pt, bottom = 1pt,
    boxrule = 0pt, bottomrule = 1pt, toprule = 1pt}

\title{Com$^2$: A Causal-Guided Benchmark for Exploring Complex Commonsense Reasoning in Large Language Models}

\author{\normalsize{Kai Xiong$^\spadesuit$}\quad
\normalsize{Xiao Ding$^\spadesuit$\footnotemark[2]}\quad
\normalsize{Yixin Cao$^\clubsuit$}\quad
\normalsize{Yuxiong Yan$^\spadesuit$}\quad
\normalsize{Li Du$^\blacklozenge$}\quad
\normalsize{Yufei Zhang$^\spadesuit$}\quad\\[-.05cm]
\normalsize{\textbf{Jinglong Gao}$^\spadesuit$}\quad
\normalsize{\textbf{Jiaqian Liu}$^\blacktriangle$}\quad
\normalsize{\textbf{Bing Qin}$^\spadesuit$}\quad
\normalsize{\textbf{Ting Liu}$^\spadesuit$}\\[-.05cm]
{\fontsize{10pt}{12pt}\selectfont $^\spadesuit$Research Center for Social Computing and Interactive Robotics}\\[-.06cm]{\fontsize{10pt}{12pt}\selectfont Harbin Institute of Technology, Harbin, China}\\[-.06cm]
\fontsize{10pt}{12pt}\selectfont{$^\clubsuit$Institute of Trustworthy Embodied AI, Fudan University, Shanghai, China}\\[-.06cm]
\fontsize{10pt}{12pt}\selectfont{$^\blacklozenge$Beijing Academy of Artificial Intelligence, Beijing, China}\\[-.06cm]
\fontsize{10pt}{12pt}\selectfont{$^\blacktriangle$Nanjing University, Nanjing, China}\\[-.1cm]
{\small\texttt{\{kxiong, xding, yxyan, yfzhang, jlgao, qinb, tliu\}@ir.hit.edu.cn}}\\[-.15cm]
{\small\texttt{yxcao@fudan.edu.cn}}
}

\begin{document}
\maketitle
\renewcommand{\thefootnote}{\fnsymbol{footnote}}
\footnotetext[2]{Corresponding Author}

\begin{abstract}
Large language models~(LLMs) have mastered abundant simple and explicit commonsense knowledge through pre-training, enabling them to achieve human-like performance in simple commonsense reasoning. Nevertheless, LLMs struggle to reason with complex and implicit commonsense knowledge that is derived from simple ones~(such as understanding the long-term effects of certain events), an aspect humans tend to focus on more. Existing works focus on complex tasks like math and code, while complex commonsense reasoning remains underexplored due to its uncertainty and lack of structure. To fill this gap and align with real-world concerns, we propose a benchmark Com$^2$ focusing on complex commonsense reasoning. We first incorporate causal event graphs to serve as structured complex commonsense. Then we adopt causal theory~(e.g., intervention) to modify the causal event graphs and obtain different scenarios that meet human concerns. Finally, an LLM is employed to synthesize examples with slow thinking, which is guided by the logical relationships in the modified causal graphs. Furthermore, we use detective stories to construct a more challenging subset. Experiments show that LLMs struggle in reasoning depth and breadth, while post-training and slow thinking can alleviate this. The code and data are available at \href{https://github.com/Waste-Wood/Com2}{Com$^2$}.
\end{abstract}

\section{Introduction}



Large Language Models~(LLMs) have made significant advances in various fields~\cite{plaat2024reasoning,ying2024intuitive}, demonstrating strong generalization and reasoning capabilities. Through massive pre-training, LLMs~\cite{liu2024deepseek,yang2024qwen2} can learn and extract knowledge from large amounts of unlabeled text.
Recent iterations, such as OpenAI's o1~\cite{o1} and Deepseek R1~\cite{liu2024deepseek}, have further enhanced these models' reasoning abilities by implementing test-time scaling strategies. These strategies encourage longer Chain-of-Thought~(CoT) to handle more complex queries, mimicking human-like reflection and correction mechanisms to improve reasoning. While these improvements have focused largely on tasks related to math~\cite{aime_problems_and_solutions} and code~\cite{quan2025codeelo}, the realm of complex commonsense reasoning remains relatively unexplored.

\begin{figure}[t]
    \centering
    \includegraphics[width=0.94\linewidth]{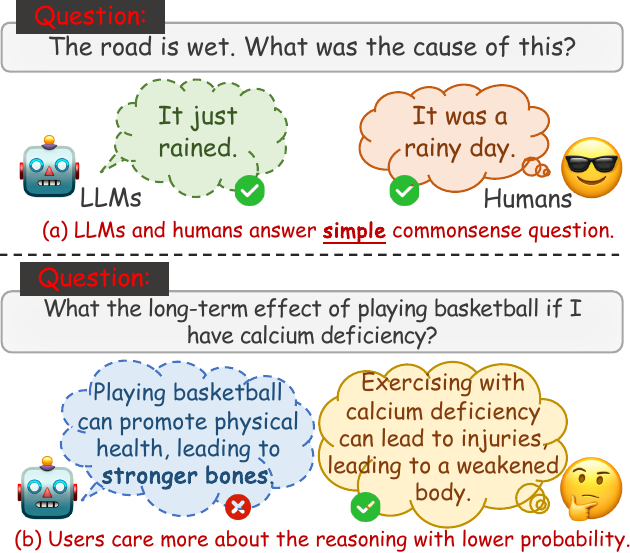}
    \caption{(a)~An one-step reasoning question. (b)~Users care more about uncommon outcomes.}
    \label{fig:intro}
    \vspace{-0.2cm}
\end{figure}

In this paper, we introduce Com$^2$, a benchmark designed specifically to assess the complex commonsense reasoning in advanced LLMs. We argue that, despite their impressive performance on existing benchmarks~\cite{wei2022chain, xiong2024meaningful}, LLMs are not inherently adept at handling complex commonsense reasoning. In real-world scenarios, humans frequently solve complex problems that involve long, multifaceted chains of commonsense reasoning. These problems often require integrating simple and explicit commonsense concepts, sometimes with cognitive biases or thinking traps. 
As shown in Figure~\ref{fig:intro}~(a), an example of current one-step commonsense reasoning task, which can be easily answered once the relevant knowledge is known.
While, we highlight LLMs, even with the use of test-time scaling, the endogenous defect of LLMs may lead to failure in thinking traps, such as uncommon scenarios~(``playing basketball with calcium deficiency'' in Figure~\ref{fig:intro}~(b)).

The challenge of curating a complex commonsense reasoning dataset stems from two primary factors. First, unlike math or code, the expression of commonsense knowledge is often informal and context-dependent. While mathematical or programming tasks typically have clear and formalized rules, commonsense reasoning is more nuanced and susceptible to interpretation, which complicates creating structured datasets. Second, commonsense reasoning tasks rarely have a universally accepted ground truth, which makes datasets ambiguous or difficult to use. This lack of clear, objective answers can lead to confusion and inconsistencies in commonsense reasoning benchmarks.

To address the above challenges, we first adopt causal event graphs~(CEGs)~\cite{ding2019elg,du2021excar} as a representation of complex commonsense knowledge, as CEGs encode complex and logically rigorous relationships among events. Based on CEGs, the most likely outcome is treated as the assumed truth. Hereafter, we use causal theory~\cite{pearl2018book} to apply modifications such as intervention on CEGs to construct different commonsense reasoning scenarios, which meet human concerns. Finally, we synthesize reasoning tasks~(Com$^2$-main) of direct, decision, transition, intervention, and counterfactual based on the modified CEGs. Apart from Com$^2$-main, we additionally propose a more challenging set Com$^2$-hard, which is based on detective stories and consists of decision, intervention, and counterfactual tasks.

Based on Com$^2$, we first evaluate a wide range of existing LLMs and then conduct several in-depth analyses. We have the following key findings and insights: 
(1)~commonsense reasoning remains a significant challenge for LLMs, as even with long thinking chains, performance does not always improve and may even degrade;
(2)~LLMs possess acceptable counterfactual reasoning abilities after massive pre-training;
(3)~LLMs have limited reasoning breadth, which causes them to struggle with uncommon or sudden scenarios. 
(4)~equipping LLMs with post-training or slow thinking can alleviate the above limitations;
Our main contributions can be summarized as follows:
\begin{itemize}
    \item We are among the first to propose a benchmark Com$^2$ which focuses on complex commonsense reasoning domains.
    \item We propose to use causal event graphs and causal theory as the backbone to create complex questions to meet real-world concerns.
    \item We comprehensively evaluate existing LLMs to gain insights for developing more capable LLMs on complex commonsense reasoning. 
\end{itemize}

\begin{figure*}[t]
    \centering
    \includegraphics[width=1\linewidth]{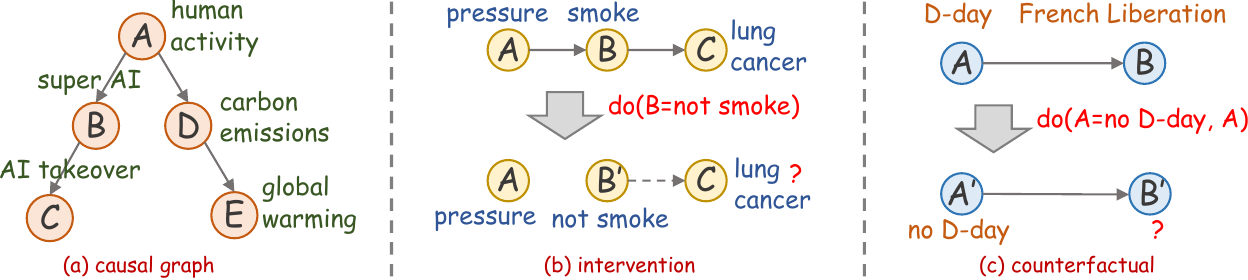}
    \caption{Three examples of (a) causal graph, (b) intervention, and (c) counterfactual, respectively.}
    \label{fig:background}
\end{figure*}

\section{Background}

\subsection{Causal Event Graph}
Causal event graph~(CEG)~\cite{ding2019elg,heindorf2020causenet} is a directed acyclic graph denoted as $\mathcal{G}=\{\mathcal{V},\mathcal{E}\}$. Figure~\ref{fig:background}~(a) shows an example of CEG, where $\mathcal{V}$ is a set of nodes, and each node $v_i$ represents a natural language event such as ``accident''. $\mathcal{E}$ is a set of directed edges, and each edge $(v_i,v_j)$ indicates a causal relationship from $v_i$ to $v_j$, which means $v_i$ is the cause and $v_j$ is the effect.

CEGs are semi-structured, which entail rigorous logical relationships among events. A CEG can represent a piece of complex and implicit commonsense knowledge, as this knowledge usually spans multiple documents and extends over time, making it difficult to directly express in text.

\subsection{Causal Theory}
We adopt causal theory~\cite{pearl2018book} to create different scenarios to match the concerns of users. We introduce two operations:

\textbf{Intervention} is an external action that actively manipulates a variable $X$ to observe its causal effect on an outcome $Y$, denoted as $do(X)$, breaking the natural dependencies of $X$ on its prior causes. Figure~\ref{fig:background}~(b) shows an intervention on ``smoke''. With intervention, we can influence common situations and steer events in a less typical direction.

\textbf{Counterfactual} refers to a hypothetical scenario that explores what would have happened if a particular past event had been different, typically denoted as $Y_X$, where $X$ is a variable that did not occur in reality. Figure~\ref{fig:background}~(c) shows a counterfactual, where ``D-day'' is an event that has occurred. Counterfactuals allow us to create hypothetical scenarios.

\section{Benchmark: Com$^2$}
\label{sec:com2}
\begin{figure*}[t]
    \centering
    \includegraphics[width=\linewidth]{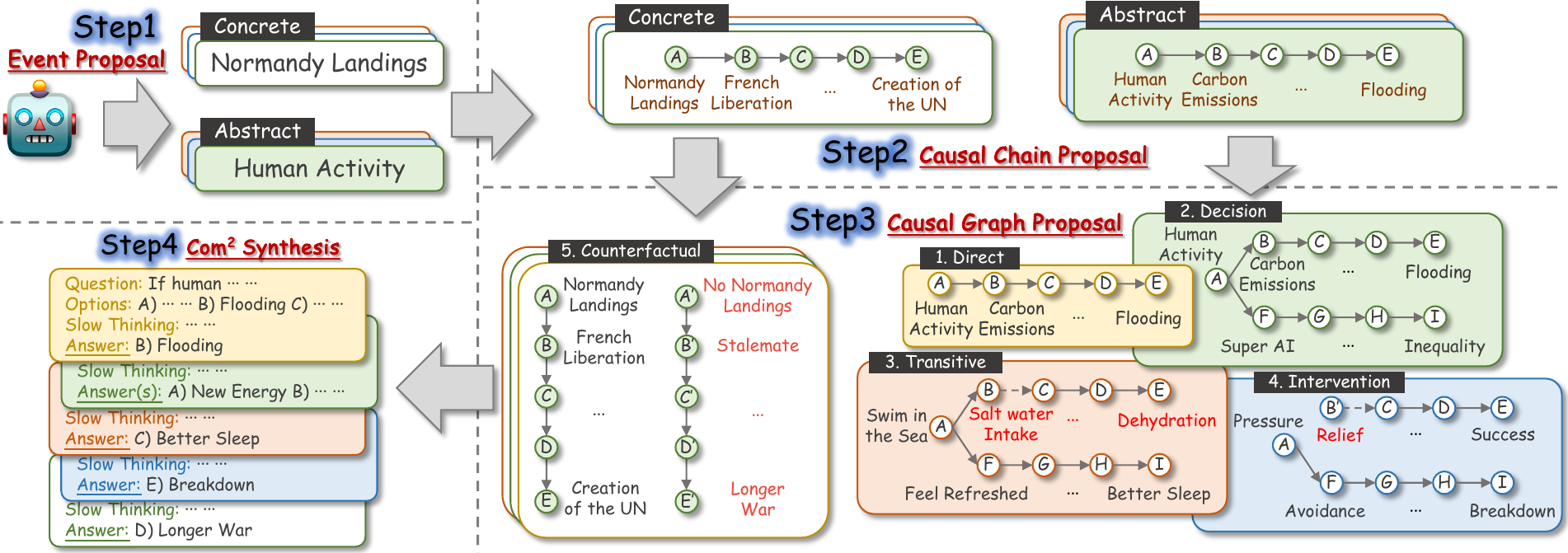}
    \caption{The whole workflow of Com$^2$ construction: (1)~Event Proposal employs LLMs to propose concrete and abstract events; (2)~Causal Chain Proposal synthesizes causal chains based on the proposed events; (3)~Causal Graph Proposal adopts causal theory to generate various causal graphs based the synthesized causal chains; (4)~Com$^2$ Synthesis creates multi-choice and multi-select questions with slow thinking, which is guided by the causal graphs.}
    \label{fig:model}
\end{figure*}

To bridge the gap between user concerns and model capability in complex commonsense reasoning, we propose Com$^2$ benchmark. As shown in Figure~\ref{fig:model}, its creation involves 4 steps: (1)~\textbf{Event Proposal} proposes concrete and abstract events for counterfactual and other scenarios, respectively. (2)~\textbf{Causal Chain Proposal} constructs causal chains to build simple scenarios. (3)~\textbf{Causal Graph Proposal} leverages causal theory to synthesize diverse causal graphs, building complex scenarios to meet user concerns. (4)~\textbf{Com$^2$ Synthesis} generates multi-choice and multi-select examples based on the causal graphs to build the benchmark. Refer to Appendix~\ref{app:com2_prompt} for all prompts.

\subsection{Event Proposal}
To construct causal graphs, we first propose some events as the seeds for generation. Specifically, we collect $k$ diverse events from an existing causal event graph~\cite{heindorf2020causenet}, and use them as $k$-shot exemplars to prompt ChatGPT~\cite{achiam2023gpt} to generate two types of events: 

$\bullet$~\textbf{Concrete}: a concrete event $e^c_i$ is a real-world occurrence that has taken place, distinguishing it from abstract events. These events are proposed for counterfactual causal graph construction.

$\bullet$~\textbf{Abstract}: an abstract event $e^a_i$ is a generalized occurrence not bound to a specific time, place, or instance. It represents categories of actions or phenomena. Abstract events are proposed in preparation for the other types of causal graphs.

Finally, we propose $n$ concrete and $n$ abstract events, denoted as $E^c=\{e^c_1, e^c_2,\cdots,e^c_n\}$ and $E^a=\{e^a_1,e^a_2,\cdots,e^a_n\}$, respectively.

\subsection{Causal Chain Proposal}
Based on the concrete and abstract events $E^c$ and $E^a$, we create corresponding causal chains to act as the basis for constructing causal graphs. Each event in $E^c$ and $E^a$ serves as the root event of its respective causal chain. We then use carefully designed prompts with ChatGPT~\cite{achiam2023gpt} to generate concrete and abstract causal chains of 5 events. Hence, each concrete or abstract causal chain represents a specific simple scenario.

Finally, we obtain $n$ concrete and $n$ abstract causal chains, denoted as $S^c=\{s_1^c,s^c_2,\cdots,s^c_n\}$ and $S^a=\{s_1^a,s^a_2,\cdots,s^a_n\}$, respectively.

\subsection{Causal Graph Proposal}
To align with the concerns of users, we adopt the causal theory~\cite{pearl2018book,xiong2022reco} and then create 5 kinds of causal graphs, which are demonstrated as follows:

$\bullet$~\textbf{Direct}: based on each abstract causal chain $s^a_i$~(Direct in Step3), the direct causal graph mirrors $s^a_i$. This reflects the long-term impact of a certain event, which is the simplest scenario.

$\bullet$~\textbf{Decision}: for each abstract causal chain $s^a_i$, we employ a prompt to generate another causal chain with undesirable outcomes~(Decision in Step3). The new chain shares the same root event as $s^c_i$, forming a decision graph with $s^a_i$. This represents a complex scenario leading to negative outcomes.

$\bullet$~\textbf{Transition}: for each abstract causal chain $s^a_i$, we use a prompt to generate another causal chain with causal transitive problems like scene drift~\cite{xiong2022reco}~(Transition in Step3). As the causal chain or the reasoning depth increases, the plausibility may gradually diminish. This transition graph simulates such complex scenarios.

$\bullet$~\textbf{Intervention}: based on each abstract causal chain $s^a_i$, we intervene in $s^a_i$ to cut off the causal transmission from the root event to the tail event. Next, we employ a prompt to generate another causal chain with a lower probability of occurring than $s^a_i$ to form the intervention graph~(Intervention in Step3). This intervention graph shows a complex scenario where a sudden event occurs and causes the causal chain to take an uncommon direction.

$\bullet$~\textbf{Counterfactual}: for each concrete causal chain $s^c_i$, we first create a counterfactual event of the root event in $s^c_i$, and then use a prompt to create a counterfactual causal chain and form the counterfactual graph~(Counterfactual in Step3). According to the causal theory, this should represent the most challenging scenario among all types.

Finally, we acquire 5 kinds of causal graphs, each representing a complex scenario. They could be used to create complex commonsense reasoning problems to align with the concerns of users.

\subsection{Com$^2$-main Synthesis}
At this step, we synthesize Com$^2$-main with 5 tasks featuring complex commonsense reasoning questions based on the causal graphs. Specifically:

$\bullet$~\textbf{Direct}: based on each direct CEG, we use the head event to formulate a multiple-choice~(MCQ) to ask for a long-term outcome, with the tail event as the answer. Distractors are added as alternative options. This task aligns with the concern of users about the long-term impact of events.

$\bullet$~\textbf{Decision}: for each decision CEG, we employ the root event to create a multi-select question~(MSQ) to seek suggestions for preventing unacceptable events. Valid interventions serve as correct answers, with distractors as alternatives. This task aligns with the user about seeking advice.

$\bullet$~\textbf{Transition}: for each transition CEG, we create an MCQ with the root event to ask for a long-term outcome. The tail event in the causal chain with transitive problems is a distractor, while the tail event in the reasonable chain is the answer, with additional distractors as alternatives. This measures the reliability of LLMs as reasoning depth grows.

$\bullet$~\textbf{Intervention}: for each intervention CEG, we create an MCQ with the root event to ask for a long-term outcome. The intervention is an additional event added to the problem. The tail event in the intervened causal chain serves as a distractor, and the tail event in the reasonable causal chain is the answer. Additional distractors are included as alternatives. This task aligns with the user concern about the outcome when sudden events occur.

$\bullet$~\textbf{Counterfactual}: for each counterfactual CEG, we use the concrete causal chain and the counterfactual event to create an MCQ to ask for a long-term outcome. Some distractors are included as alternatives. This aligns with the user concern about long-term outcomes in hypothetical scenarios.

Finally, for each example in each task, we additionally provide a slow thinking, which is guided by corresponding CEG and consists of systematic analysis, divide-and-conquer, self-refinement, and context identification~\cite{wu2024comparative}. Hence, the above 5 tasks form Com$^2$-main.

\subsection{Com$^2$-hard Synthesis}
Besides Com$^2$-main, we aim to create a more challenging version of Com$^2$. Since detective stories feature multiple interwoven clues, requiring complex combinations of reasoning to deduce the conclusion. This leads to a far more complex scenario than previous CEGs. Inspired by this, we develop Com$^2$-hard based on BMDS~\cite{BMDS}. BMDS is a collection of over 400 stories comprising annotations for the key clue, and evidence. Specifically, we first provide ChatGPT with the detective story to extract the \texttt{criminal(s)}. Next, ChatGPT summarizes all the facts and actions of all individuals involved in the story~(denoted as \texttt{clues}). Finally, we create 3 tasks:

$\bullet$~\textbf{Decision}: we create an MSQ to ask for suggestions to prevent the crime from occurring. All plausible interventions to the crime are treated as answers. Some distractors will be incorporated to form options. The slow thinking is also guided by the case-solving route in the detective story, and has the same actions as each Com$^2$-main task.

$\bullet$~\textbf{Intervention}: we exclude the key clue in the \texttt{clues}, and then create a MCQ to ask for the criminal(s). The key clue serves as an intervention added to the question. The \texttt{criminal(s)} is treated as the answer. Some suspects will be incorporated to form options. The slow thinking is guided by the case-solving route in the detective story, and has the same actions as each Com$^2$-main task.

$\bullet$~\textbf{Counterfactual}: we first create a counterfactual based on the detective story, and then create a MCQ to ask for the outcome of a counterfactual. The counterfactual outcome serves as the answer. Some distractors are incorporated to form options. The slow thinking is guided by the case-solving route and the counterfactual, which has the same actions as each Com$^2$-main task.

\subsection{Com$^2$ Statistics}

\begin{table}[t]
\small

\centering
\resizebox{\linewidth}{!}{
\begin{tabular}{lcccccc}
\toprule
\textbf{Com$^2$} & \textbf{Dir.} & \textbf{Dec.} & \textbf{Trans.} & \textbf{Inter.} & \textbf{Counter.} & \textbf{Total}\\ \midrule
\textbf{Main} & 500 & 500 & 500 & 500 & 500 & 2,500\\
\textbf{Hard} & - & 418 & - & 418 & 418 & 1,254\\ \bottomrule
\end{tabular}}
\caption{Statistics of Com$^2$ benchmark}
\label{tab:com2}
\vspace{-0.2cm}
\end{table}

We conduct the above process to create Com$^2$, the API of ChatGPT we used is \texttt{gpt-4o-mini}. As shown in Table~\ref{tab:com2}, there are 500 examples for each task in Com$^2$-main, and 418 examples for each task in Com$^2$-hard. Examples can refer to Appendix~\ref{app:examples}.

\subsection{Human Evaluation}
We conduct a human evaluation to evaluate the quality of Com$^2$. The details~(such as the pay and agreements) of the human evaluation can be referred to Appendix~\ref{app:human_eval}. By providing human annotators with the causal graphs, humans can achieve average accuracies of 92\% and 90\% for Com$^2$-main and Com$^2$-hard, respectively. The reliability of the slow thinking process is 88\%. These demonstrate the quality of Com$^2$ is satisfactory.

\begin{table*}[t]
\small
\centering
\resizebox{\textwidth}{!}{
\begin{tabular}{llcccccc|cccc|c}
\toprule
\multicolumn{2}{c}{\multirow{2}{*}{\textbf{LLMs}}} & \multicolumn{6}{c}{\textbf{Main}} & \multicolumn{4}{c}{\textbf{Hard}} & \multirow{2}{*}{\textbf{Overall}} \\ \cmidrule{3-12}
\multicolumn{2}{c}{} & Dir. & Dec. & Trans. & Inter. & Counter. & Avg. & Dec. & Inter. & Counter. & Avg. &  \\ \midrule
\multirow{11}{*}{\textbf{General}} & Qwen2-7B & 80.20 & 59.25 & 47.60 & \textbf{34.00} & 69.60 & 58.13 & 28.13 & \textbf{57.76} & 78.23 & 54.71 & 56.42 \\
 & Qwen2.5-7B & 83.40 & \textbf{67.83} & \textbf{49.80} & 32.80 & \textbf{73.40} & \textbf{61.42} & 31.13 & 51.07 & 74.88 & 52.36 & 56.89 \\
 & Qwen2.5-14B & 80.40 & 66.95 & 48.20 & 31.80 & 72.00 & 59.84 & 34.94 & 51.79 & 75.84 & 54.19 & 57.02 \\
 & Qwen2.5-32B & \textbf{83.60} & 65.16 & 48.80 & 33.80 & 72.40 & 60.73 & 30.21 & 54.89 & \textbf{79.19} & 54.80 & 57.77 \\
 & LLaMA-3.1-8B & 83.20 & 58.04 & 47.00 & 30.40 & 71.40 & 58.01 & 37.62 & 48.93 & 74.16 & 53.56 & 55.79 \\
 & LLaMA-3.2-1B & 68.20 & 27.16 & 35.60 & 24.20 & 47.20 & 40.52 & 3.14 & 27.92 & 28.95 & 20.01 & 30.27 \\
 & LLaMA-3.2-3B & 81.20 & 58.04 & 40.20 & 29.20 & 72.40 & 56.20 & 21.22 & 48.45 & 62.20 & 43.96 & 50.08 \\
 & Gemma2-9B & 78.20 & 11.99 & 45.20 & 26.40 & 68.40 & 46.16 & 43.20 & 53.46 & 77.75 & 58.13 & 52.15 \\
 & Gemma2-27B & 77.40 & 60.29 & 49.20 & 28.40 & 69.60 & 56.97 & 45.73 & 49.64 & 75.36 & 56.90 & 56.94 \\
 & GPT-4o-mini & 83.20 & 62.54 & 49.20 & 31.40 & 71.20 & 59.50 & 33.46 & 53.46 & 78.95 & 55.29 & 57.40 \\
 & GPT-4o & 80.60 & 66.43 & 48.40 & 32.20 & 68.80 & 59.26 & 45.10 & 56.09 & 77.99 & 59.72 & 59.49 \\ \midrule
\multirow{5}{*}{\textbf{Reasoning}} & Open-O1 & 75.60 & 41.67 & 43.80 & 30.40 & 60.00 & 50.29 & 62.47 & 52.03 & 71.05 & 61.84 & 56.07 \\
 & Marco-o1 & 77.60 & 41.77 & 43.60 & 31.80 & 65.80 & 52.11 & \textbf{62.64} & 53.94 & 71.77 & \textbf{62.78} & 57.45 \\
 & R1-distilled & 75.20 & 56.51 & 43.40 & 30.00 & 68.20 & 54.65 & 60.22 & 54.24 & 73.64 & 62.70 & \textbf{58.68} \\
 & QwQ-32B & 79.80 & 59.82 & 47.40 & 32.00 & 64.60 & 56.70 & 41.81 & 44.39 & 69.86 & 52.01 & 54.36 \\
 & o1-mini & 80.00 & 32.64 & 47.80 & 30.00 & 66.60 & 51.48 & 43.44 & 51.79 & 74.40 & 56.54 & 54.01 \\
 \bottomrule
\end{tabular}}
\caption{The overall performance of various LLMs on Com$^2$. Dir., Dec., Trans., Inter., and Counter. represent Direct, Decision, Transition, Intervention, and Counterfactual, respectively. ``Avg.'' denotes the average performance across all tasks in Com$^2$-main or Com$^2$-hard. ``Total'' indicates the overall performance on Com$^2$.}
\label{tab:main_results}
\end{table*}

\section{Experiments}
\subsection{Investigated LLMs}
Based on Com$^2$, we adopt a wide range of different kinds of LLMs to evaluate the complex commonsense reasoning capabilities of them:

$\bullet$ \textbf{General LLMs}: (1)~Qwen~\cite{yang2024qwen2}: 7B Qwen2-Instruct. 7B, 14B, and 32B Qwen2.5-Instruct. (2)~LLaMA~\cite{dubey2024llama}: LLaMA-3.1-8B-Instruct. 1B and 3B LLaMA-3.2-Instruct. (3)~Gemma~\cite{team2024gemma}: 9B and 27B Gemma2-it. (4)~ChatGPT~\cite{achiam2023gpt}: GPT-4o and GPT-4o-mini.

$\bullet$ \textbf{Reasoning LLMs}: (1)~Open-O1~\cite{Open-O1}, which is a reasoning-tunned LLM based on LLaMA-3.1-8B-Instruct. (2)~Marco-o1~\cite{zhao2024marco}, which is a reasoning-tunned LLM based on Qwen2-7B-Instruct. (3)~R1-distilled~\cite{liu2024deepseek}, which is a reasoning LLMs based on LLaMA-3.1-8B and from DeepSeek-R1. (4)~QwQ-32B-Preview~\cite{yang2024qwen2}, which is a complex reasoning LLM based on Qwen2.5-32B. (5)~o1-mini, which is a powerful reasoning LLM developed by OpenAI.

\subsection{Evaluation Details}
For each task in Com$^2$, we adopt accuracy to quantify the performance of each LLM. Specifically, for decision task that consists of multi-select questions, we design a soft strategy. For each multi-select question, the accuracy is determined by the proportion of correctly predicted answer choices in the full set of correct answers. A perfect match scores 1, partial matches are scored based on the fraction of correct choices predicted, predictions with wrong choices score 0.

We follow a zero-shot setting to evaluate each LLM. Each LLM is asked to provide a CoT~\cite{wei2022chain} first and then give the answer. All evaluation prompts can refer to Appendix~\ref{app:eval}.

\subsection{Overall Results}

The overall results are shown in Table~\ref{tab:main_results}, from which we can have the following observations:

(1)~For nearly all LLMs, the average performance on Com$^2$-main is higher than that on Com$^2$-hard. This is because Com$^2$-hard features multiple interwoven clues, requiring different and complex combinations of reasoning to reach conclusion.

(2)~According to the causal theory~\cite{pearl2018book}, Counterfactual should be the most difficult task. Interestingly, LLMs perform better on Counterfactual than on Decision, Transition, and Intervention tasks. We suppose LLMs might obtain adequate hypothetical reasoning capability through massive pre-training.

(3)~Almost all LLMs perform better on Direct than on the other tasks, as this might come from the fact that LLMs just need to reason through a causal chain only. LLMs might perform satisfactorily in terms of reasoning depth.

(4)~On Com$^2$-main, Transition and Intervention are much harder than the others. This reveals that LLMs might possess an issue with insufficient handling of long causal dependencies when reasoning depth increases. Furthermore, LLMs find it difficult to handle sudden events~(intervention). This demonstrate LLMs have limited reasoning breadth.

(5)~On Com$^2$-main, Intervention is harder than Decision, but Com$^2$-hard, the situation is completely reversed. The main reason is that LLMs might have memorized the detective through pre-training, making it easier to deduce the criminal(s).

(6)~On Com$^2$-hard and overall performance, limited access LLMs~(GPT-4o, GPT-4o-mini) have some advantages over open access LLMs. It is due to the better general ability of ChatGPT.

(7)~o1-mini and QwQ do not perform as well as other reasoning LLMs, which might be due to their overfitting on the other domains.

(8)~Larger LLMs do not necessarily lead to better performance~(Qwen2.5 series). The complex commonsense reasoning in LLMs may also require a certain scale to emerge~(LLaMA-3.2 series).

\subsection{General LLMs v.s. Reasoning LLMs}
We compared reasoning LLMs to their corresponding general LLMs~(e.g., Open-O1 and LLaMA-3.1-8B-Instruct), and draw the following conclusions:

(1)~On Com$^2$-hard, reasoning-tunned LLMs are superior to their corresponding general LLMs~(such as Open-O1 and LLaMA-3.1-8B-Instruct), as the massive thought process could help LLMs to understand and deal with more complex scenarios.

(2)~However, on Com$^2$-main, reasoning LLMs usually do not gain advantages over the general LLMs, and their performance decreases. This might result from the overthinking of reasoning-tunned LLMs. It also indicates that reasoning LLMs also require the appropriate level of problem difficulty to match their capabilities.

(3)~On Com$^2$-hard, the improvement of reasoning LLMs compared to general models is mainly in Decision task. It is also the hardest task in Com$^2$-hard. It further proves that the more difficult the task, the greater advantage reasoning LLMs have.

\section{Analysis}
To further study the complex commonsense reasoning scenarios, we design several in-depth analyses. (1)~We train LLMs on complex commonsense reasoning examples to investigate their effectiveness of them. (2)~We analyze the relationship between model performance and output token count to discuss the scaling law during inference. (3)~We investigate the effectiveness of the slow thinking within each example in Com$^2$ by providing it to LLMs for complex commonsense reasoning~(Com$^2$).

\subsection{Effect of Post-training}
To investigate whether complex commonsense reasoning data can help LLMs to improve themselves, we employ the Com$^2$-main construction process~(Sec.~\ref{sec:com2}) to construct a training dataset. The training data consists of 8,386 examples, which has 4184, 476, 475, 2757, and 
476 questions for tasks of Direct, Decision, Transition, Interevntion, and Counterfactual, respectively. We choose LLaMA-3.1-8B-Instruct~(LLaMA) and Qwen2-7B-Instruct~(Qwen2) for training. Training details and cases can refer to Appendix~\ref{app:training} and~\ref{app:case}, respectively.

\begin{figure}
    \centering
    \includegraphics[width=1\linewidth]{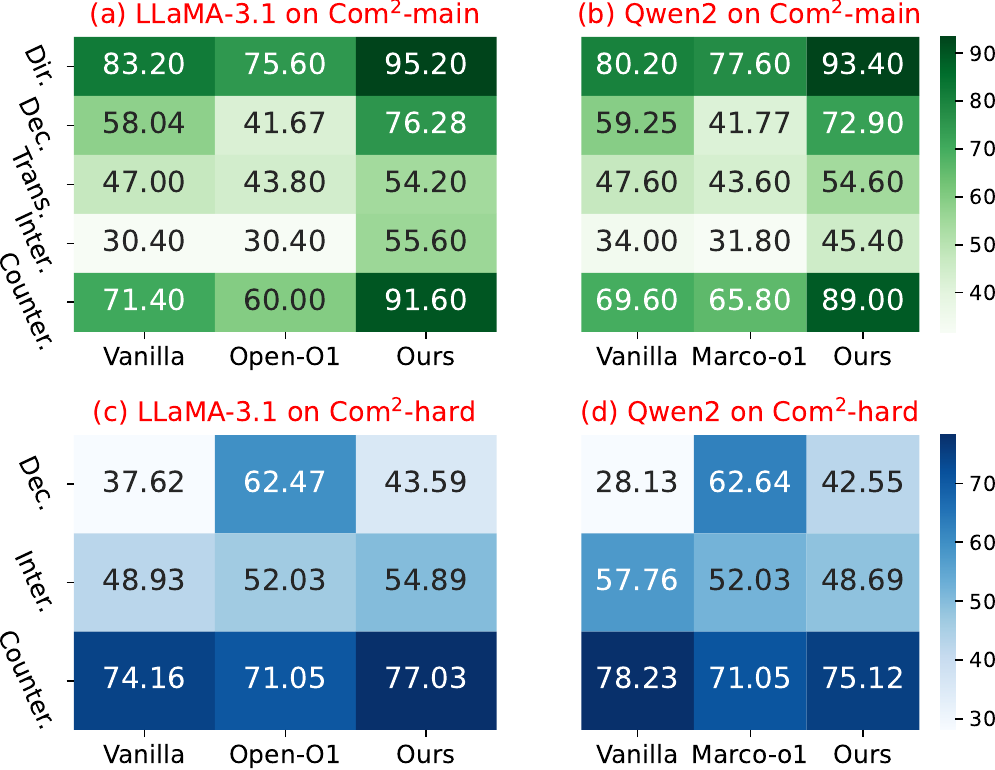}
    \caption{Overall performance of trained LLMs~(Ours) and baselines on Com$^2$ benchmark.}
    \label{fig:analysis1}
\end{figure}

We evaluate the trained LLMs~(denoted as Ours) and baselines on Com$^2$. Note that Com$^2$-hard is treated as an out-of-distribution~(OOD) dataset. The overall results can refer to Figure~\ref{fig:analysis1}, from which we have the following observations:

(1)~After training, LLaMA and Qwen2 can obtain significant improvement over vanilla LLMs on Com$^2$-main. The improvement over vanilla LLMs on Com$^2$-hard is noticeable, but relatively much smaller. This is mainly because Com$^2$-hard is OOD. It also indicates that LLMs can learn complex commonsense reasoning abilities on simpler tasks and transfer them to more complex tasks.

(2)~On Com$^2$-main, the improvement of trained LLMs is much smaller on Transition task than on the other tasks. We suppose more examples are required to teach LLMs to learn the differences in quantitative and scene information.

(3)~On Com$^2$-hard, the improvement of trained LLMs is substantial on Decision task. The main reason might be that LLMs cannot deal with multi-select questions well before training.

(4)~On Com$^2$-hard, the trained LLMs perform worse than reasoning LLMs~(Open-O1 or Marco-o1). This reveals that more types of training data are required for improvement.







\begin{figure}[t]
    \centering
    \includegraphics[width=0.95\linewidth]{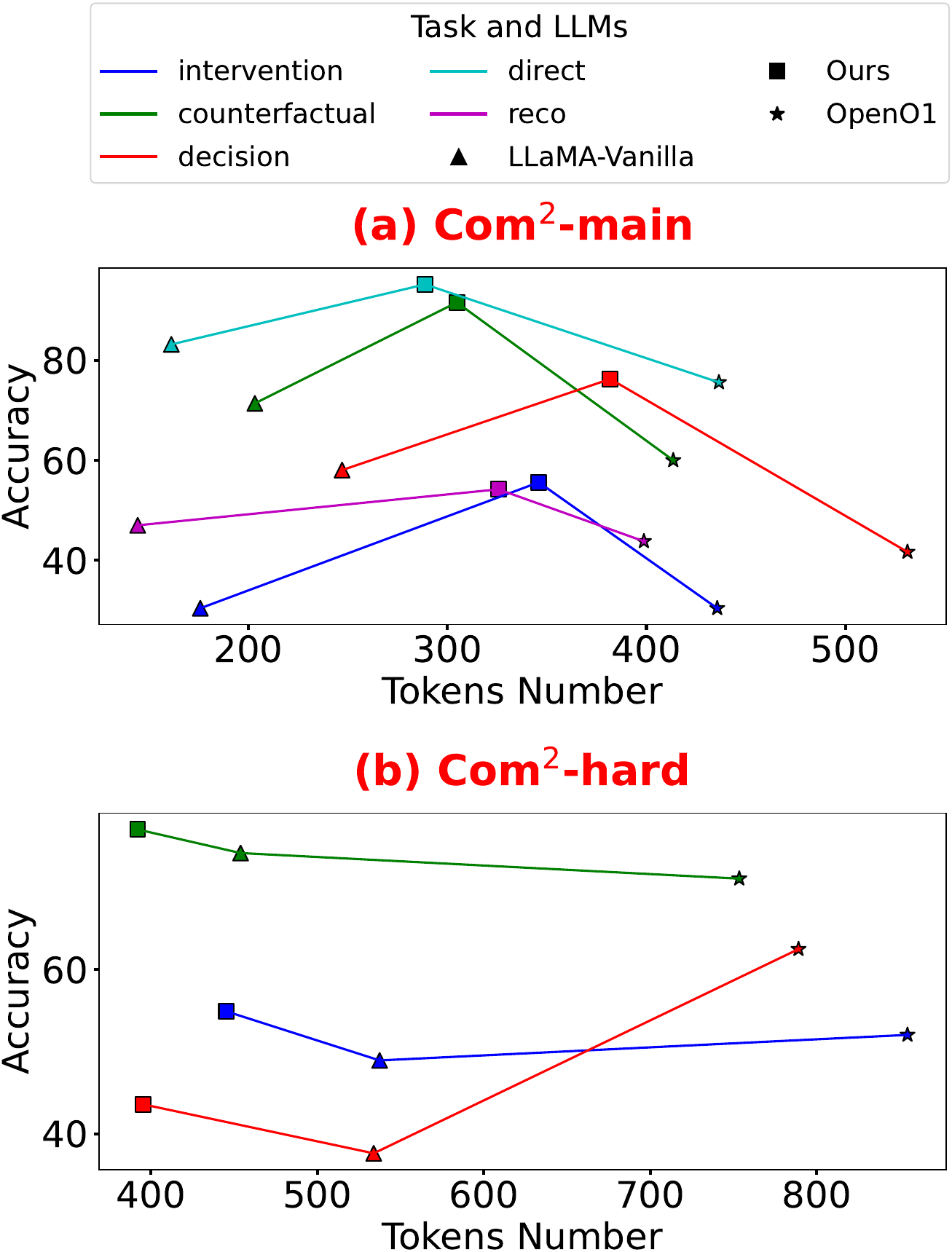}
    \caption{The relationships between LLMs performance and output token count.}
    \label{fig:analysis4}
\end{figure}

\subsection{Test-time Scaling Law}
To discuss the scaling law during inference, we statistic the number of tokens output by LLMs. Overall results are shown in Figure~\ref{fig:analysis4}, we can infer:

(1)~Outputting more tokens does not always improve performance. The test-time scaling law in math and code may not be effective for commonsense reasoning, as they can still fall into common sense shortcuts. Post-training can alleviate this to improve performance and reduce token usage.

(2)~The conclusion is also supported by the performance on Intervention~(Figure~\ref{fig:analysis1}), as reasoning LLMs perform much worse than our method.

(3)~The overall token usage is still much smaller compared to math. It is worth looking forward to the test-time scaling in commonsense reasoning.





\begin{figure}
    \centering
    \includegraphics[width=0.95\linewidth]{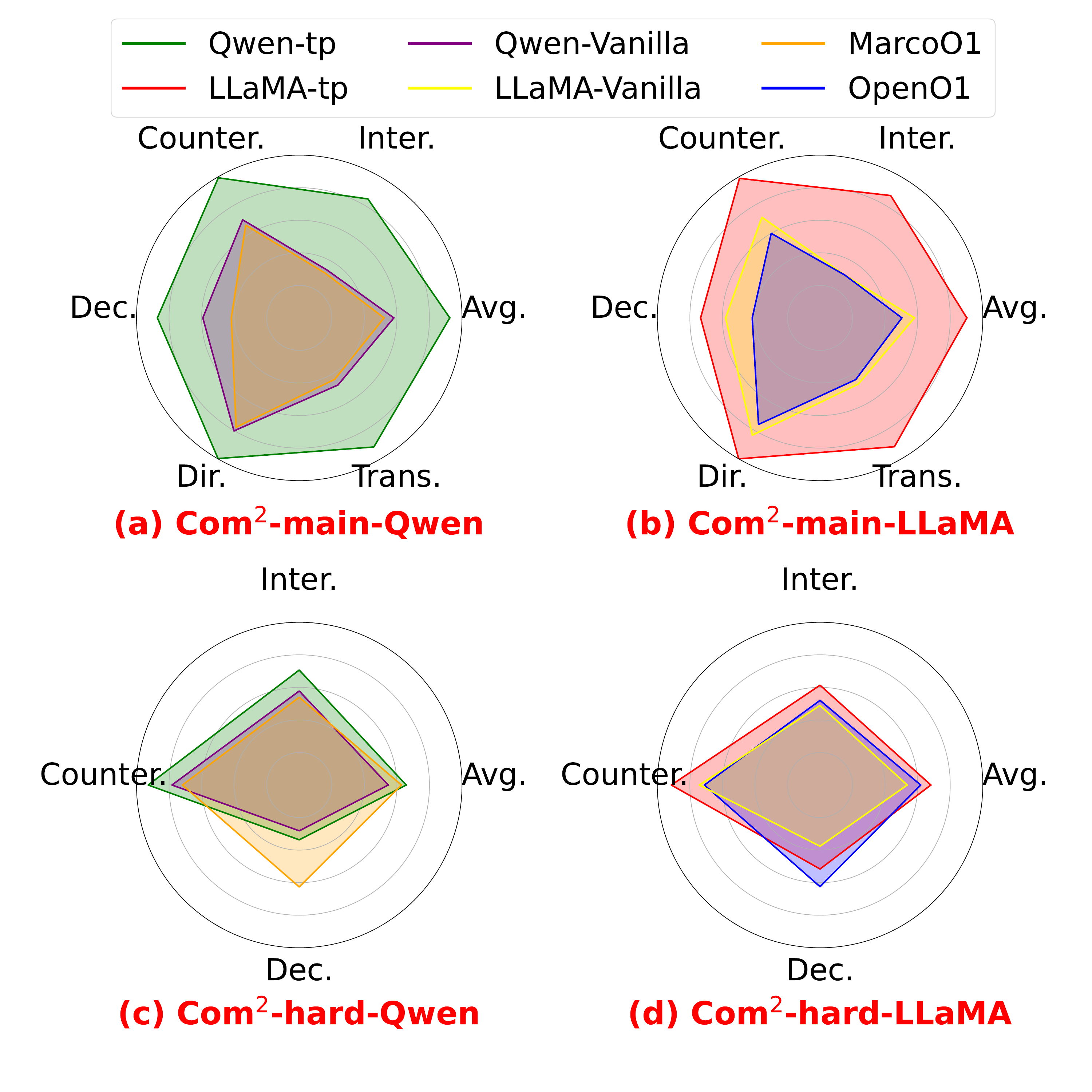}
    \caption{The overall performance of LLaMA and Qwen guided by slow thinking process, as well as baselines.}
    \label{fig:analysis3}
\end{figure}

\subsection{Effect of the Slowing Thinking}
To investigate whether the slow thinking in each example of Com$^2$ can help LLMs to conduct complex commonsense reasoning, we provide LLMs with the slow thinking for answer selection. We choose Qwen2-7B-Instruct and LLaMA-3.1-8B-Instruct for experiments. The LLMs promoted with slow thinking are denoted as Qwen-tp and LLaMA-3.1-tp. The overall performance of Qwen-tp, LLaMA-tp, and baselines can refer to Figure~\ref{fig:analysis3}. We can draw the following conclusions:

(1)~After providing LLMs with the slow thinking, both LLMs achieved significant improvement on both Com$^2$-main and Com$^2$-hard. The reason is the slow thinking can provide a detailed analysis of the question and help to clarify the causal relationships among events, leading to a boost in performance.

(2)~While for Decision task, it has not been elevated to an extremely high level. We suppose it is due to the multi-select nature of Decision task, and the slow thinking is too long for LLMs to extract key information for reasoning.

\section{Related Work}
\subsection{Reasoning in LLMs}
LLMs have revolutionized the paradigm of using deep neural networks for reasoning. Employing LLMs for reasoning can be categorized into prompt-based and post-training-based methods.

As for prompt-based methods, the emergence of GPT-3~\cite{brown2020language,chen2021evaluating} opened the door for LLMs to perform zero-shot and few-shot reasoning. \citet{wei2022chain} proposed chain-of-thought~(CoT), which elicit reasoning in LLMs and obtain self-interpretability. \citet{kojima2022large} and \citet{zhangautomatic} respectively proposed zero-shot and auto CoT to overcome the need for human annotation. \citet{fu2022complexity} and \citet{yao2024tree} respectively designed complex CoT and tree-of-thought to further improve the complexity and performance of CoT. Moreover, some works utilize multi-agent collaboration to improve the reasoning in LLMs~\cite{xiong2023examining,liang2023encouraging,duimproving}.

As for post-training-based methods, \citet{sanhmultitask} and \citet{weifinetuned} respectively proposed T0 and FLAN to train LLMs on massive NLP tasks, achieving superior zero-shot task generalization capabilities. T$k$-INSTRUCT~\cite{wang2022super} and Flan-PaLM~\cite{chung2024scaling} scaled up the task size for further enhancement. Orca~\cite{mukherjee2023orca} and Orca-2~\cite{mitra2023orca} used GPT-4 to synthesize data and then train small-scale LLMs to improve reasoning. \citet{xu2024wizardlm} investigated evolve instructions to enhance LLMs for reasoning. \citet{xiong2024meaningful} studied meaningful learning to advance abstract reasoning in LLMs. Furthermore, \citet{xiong2024diagnosing} diagnosed and remedied deficiencies in LLMs to achieve targeted improvement.\nocite{du2021excar,du2022care}

Our work mainly focuses on evaluating the reasoning capabilities of existing LLMs, especially on complex commonsense reasoning. This evaluation could provide insights for further improvement.

\subsection{Complex Reasoning in LLMs}
LLMs have managed good simple reasoning performance through pre-training and supervised fine-tuning. The performance of LLMs on complex tasks has been the focus of research fields.

\citet{zhouleast} proposed Leat-to-Most prompting to decompose complex questions into simple questions and achieve notable enhancement. \citet{yaoreact} designed ReAct to teach LLMs to use tools  to support reasoning. Recently with the release of o1, o3, and DeepSeek-R1~\cite{liu2024deepseek}, many work start to teach LLMs with slow-thinking abilities to solve complex tasks. Open-O1~\cite{Open-O1} used reflection data to teach LLMs to reflect on errors. \citet{zhao2024marco} and \citet{Slow_Thinking_with_LLMs_1} respectively devised Marco-o1 and STILL based on Monte Carlo Tree Search (MCTS) to distill slow-thinking data and train LLMs. \citet{yang2024qwen2} constructed QwQ-32-Preview to improve the complex math reasoning abilities of LLMs.

Our work mainly focuses on complex reasoning in the commonsense domain, while previous works mainly focus on math or the general domain. Our work is complementary to theirs.

\section{Conclusion}
In this paper, we propose to construct a complex commonsense reasoning benchmark called Com$^2$ to align with user demands. Com$^2$ is synthesized based on causal event graphs and guided by causal theory. Hereafter, we use Com$^2$ to evaluate a wide range of existing LLMs, and we find that existing LLMs still possess some issues with reasoning depth and breadth, which pose a challenge to the current LLMs. Further analyses reveal that equipping LLMs with post-training and slow thinking could alleviate the above issues.

\section*{Acknowledgements}
We would like to thank the anonymous reviewers for their constructive comments. The research in this article is supported by the New Generation Artificial Intelligence of China~(2024YFE0203700), National Natural Science Foundation of China under Grants U22B2059 and 62176079, and CCF-Zhipu Large Model Innovation Fund~(NO. CCF-Zhipu202401).

\section*{Limitations}
This paper still possesses some limitations. First, the dataset synthesis process could benefit from finer-grained and step-by-step guidance of causal event graphs. Second, the questions can be open-ended, which can be evaluated via LLMs-based evaluators. Lastly, the construction of Com$^2$-hard could be refined through a more robust application of causal graph techniques, along with deeper and more comprehensive analyses of the causal relationships at play. This would allow for a clearer understanding of the factors influencing Com$^2$-hard, leading to more accurate and insightful results.

\bibliography{custom}

\begin{thebibliography}{42}
\providecommand{\natexlab}[1]{#1}

\bibitem[{Achiam et~al.(2023)Achiam, Adler, Agarwal, Ahmad, Akkaya, Aleman, Almeida, Altenschmidt, Altman, Anadkat et~al.}]{achiam2023gpt}
Josh Achiam, Steven Adler, Sandhini Agarwal, Lama Ahmad, Ilge Akkaya, Florencia~Leoni Aleman, Diogo Almeida, Janko Altenschmidt, Sam Altman, Shyamal Anadkat, et~al. 2023.
\newblock Gpt-4 technical report.
\newblock \emph{arXiv preprint arXiv:2303.08774}.

\bibitem[{AIME()}]{aime_problems_and_solutions}
AIME.
\newblock Aime problems and solutions.

\bibitem[{Brown et~al.(2020)Brown, Mann, Ryder, Subbiah, Kaplan, Dhariwal, Neelakantan, Shyam, Sastry, Askell et~al.}]{brown2020language}
Tom Brown, Benjamin Mann, Nick Ryder, Melanie Subbiah, Jared~D Kaplan, Prafulla Dhariwal, Arvind Neelakantan, Pranav Shyam, Girish Sastry, Amanda Askell, et~al. 2020.
\newblock Language models are few-shot learners.
\newblock \emph{Advances in neural information processing systems}, 33:1877--1901.

\bibitem[{Chen et~al.(2021)Chen, Tworek, Jun, Yuan, Pinto, Kaplan, Edwards, Burda, Joseph, Brockman et~al.}]{chen2021evaluating}
Mark Chen, Jerry Tworek, Heewoo Jun, Qiming Yuan, Henrique Ponde De~Oliveira Pinto, Jared Kaplan, Harri Edwards, Yuri Burda, Nicholas Joseph, Greg Brockman, et~al. 2021.
\newblock Evaluating large language models trained on code.
\newblock \emph{arXiv preprint arXiv:2107.03374}.

\bibitem[{Chung et~al.(2024)Chung, Hou, Longpre, Zoph, Tay, Fedus, Li, Wang, Dehghani, Brahma et~al.}]{chung2024scaling}
Hyung~Won Chung, Le~Hou, Shayne Longpre, Barret Zoph, Yi~Tay, William Fedus, Yunxuan Li, Xuezhi Wang, Mostafa Dehghani, Siddhartha Brahma, et~al. 2024.
\newblock Scaling instruction-finetuned language models.
\newblock \emph{Journal of Machine Learning Research}, 25(70):1--53.

\bibitem[{Ding et~al.(2019)Ding, Li, Liu, and Liao}]{ding2019elg}
Xiao Ding, Zhongyang Li, Ting Liu, and Kuo Liao. 2019.
\newblock Elg: an event logic graph.
\newblock \emph{arXiv preprint arXiv:1907.08015}.

\bibitem[{Du et~al.(2021)Du, Ding, Xiong, Liu, and Qin}]{du2021excar}
Li~Du, Xiao Ding, Kai Xiong, Ting Liu, and Bing Qin. 2021.
\newblock Excar: Event graph knowledge enhanced explainable causal reasoning.
\newblock In \emph{Proceedings of the 59th Annual Meeting of the Association for Computational Linguistics and the 11th International Joint Conference on Natural Language Processing (Volume 1: Long Papers)}, pages 2354--2363.

\bibitem[{Du et~al.(2022)Du, Ding, Xiong, Liu, and Qin}]{du2022care}
Li~Du, Xiao Ding, Kai Xiong, Ting Liu, and Bing Qin. 2022.
\newblock e-care: a new dataset for exploring explainable causal reasoning.
\newblock In \emph{Proceedings of the 60th Annual Meeting of the Association for Computational Linguistics (Volume 1: Long Papers)}, pages 432--446.

\bibitem[{Du et~al.(2024)Du, Li, Torralba, Tenenbaum, and Mordatch}]{duimproving}
Yilun Du, Shuang Li, Antonio Torralba, Joshua~B Tenenbaum, and Igor Mordatch. 2024.
\newblock Improving factuality and reasoning in language models through multiagent debate.
\newblock In \emph{Forty-first International Conference on Machine Learning}.

\bibitem[{Dubey et~al.(2024)Dubey, Jauhri, Pandey, Kadian, Al-Dahle, Letman, Mathur, Schelten, Yang, Fan et~al.}]{dubey2024llama}
Abhimanyu Dubey, Abhinav Jauhri, Abhinav Pandey, Abhishek Kadian, Ahmad Al-Dahle, Aiesha Letman, Akhil Mathur, Alan Schelten, Amy Yang, Angela Fan, et~al. 2024.
\newblock The llama 3 herd of models.
\newblock \emph{arXiv preprint arXiv:2407.21783}.

\bibitem[{Fu et~al.(2023)Fu, Peng, Sabharwal, Clark, and Khot}]{fu2022complexity}
Yao Fu, Hao Peng, Ashish Sabharwal, Peter Clark, and Tushar Khot. 2023.
\newblock Complexity-based prompting for multi-step reasoning.
\newblock In \emph{The Eleventh International Conference on Learning Representations}.

\bibitem[{Hammond and Stern(2022)}]{BMDS}
Adam Hammond and Simon Stern. 2022.
\newblock The birth of the modern detective story (bmds) dataset.

\bibitem[{Heindorf et~al.(2020)Heindorf, Scholten, Wachsmuth, Ngonga~Ngomo, and Potthast}]{heindorf2020causenet}
Stefan Heindorf, Yan Scholten, Henning Wachsmuth, Axel-Cyrille Ngonga~Ngomo, and Martin Potthast. 2020.
\newblock Causenet: Towards a causality graph extracted from the web.
\newblock In \emph{Proceedings of the 29th ACM international conference on information \& knowledge management}, pages 3023--3030.

\bibitem[{Jiang et~al.(2024)Jiang, Chen, Min, Chen, Cheng, Wang, Tang, Sun, Deng, Zhao, Liu, Yan, Xie, Wang, and Wen}]{Slow_Thinking_with_LLMs_1}
Jinhao Jiang, Zhipeng Chen, Yingqian Min, Jie Chen, Xiaoxue Cheng, Jiapeng Wang, Yiru Tang, Haoxiang Sun, Jia Deng, Wayne~Xin Zhao, Zheng Liu, Dong Yan, Jian Xie, Zhongyuan Wang, and Ji-Rong Wen. 2024.
\newblock Enhancing llm reasoning with reward-guided tree search.
\newblock \emph{arXiv preprint arXiv:2411.11694}.

\bibitem[{Kojima et~al.(2022)Kojima, Gu, Reid, Matsuo, and Iwasawa}]{kojima2022large}
Takeshi Kojima, Shixiang~Shane Gu, Machel Reid, Yutaka Matsuo, and Yusuke Iwasawa. 2022.
\newblock Large language models are zero-shot reasoners.
\newblock \emph{Advances in neural information processing systems}, 35:22199--22213.

\bibitem[{Liang et~al.(2024)Liang, He, Jiao, Wang, Wang, Wang, Yang, Shi, and Tu}]{liang2023encouraging}
Tian Liang, Zhiwei He, Wenxiang Jiao, Xing Wang, Yan Wang, Rui Wang, Yujiu Yang, Shuming Shi, and Zhaopeng Tu. 2024.
\newblock Encouraging divergent thinking in large language models through multi-agent debate.
\newblock In \emph{Proceedings of the 2024 Conference on Empirical Methods in Natural Language Processing}, pages 17889--17904.

\bibitem[{Liu et~al.(2024)Liu, Feng, Xue, Wang, Wu, Lu, Zhao, Deng, Zhang, Ruan et~al.}]{liu2024deepseek}
Aixin Liu, Bei Feng, Bing Xue, Bingxuan Wang, Bochao Wu, Chengda Lu, Chenggang Zhao, Chengqi Deng, Chenyu Zhang, Chong Ruan, et~al. 2024.
\newblock Deepseek-v3 technical report.
\newblock \emph{arXiv preprint arXiv:2412.19437}.

\bibitem[{Mitra et~al.(2023)Mitra, Del~Corro, Mahajan, Codas, Simoes, Agarwal, Chen, Razdaibiedina, Jones, Aggarwal et~al.}]{mitra2023orca}
Arindam Mitra, Luciano Del~Corro, Shweti Mahajan, Andres Codas, Clarisse Simoes, Sahaj Agarwal, Xuxi Chen, Anastasia Razdaibiedina, Erik Jones, Kriti Aggarwal, et~al. 2023.
\newblock Orca 2: Teaching small language models how to reason.
\newblock \emph{arXiv preprint arXiv:2311.11045}.

\bibitem[{Mukherjee et~al.(2023)Mukherjee, Mitra, Jawahar, Agarwal, Palangi, and Awadallah}]{mukherjee2023orca}
Subhabrata Mukherjee, Arindam Mitra, Ganesh Jawahar, Sahaj Agarwal, Hamid Palangi, and Ahmed Awadallah. 2023.
\newblock Orca: Progressive learning from complex explanation traces of gpt-4.
\newblock \emph{arXiv preprint arXiv:2306.02707}.

\bibitem[{OpenAI(2024)}]{o1}
OpenAI. 2024.
\newblock Openai o1 system card.

\bibitem[{Pearl and Mackenzie(2018)}]{pearl2018book}
Judea Pearl and Dana Mackenzie. 2018.
\newblock \emph{The book of why: the new science of cause and effect}.
\newblock Basic books.

\bibitem[{Plaat et~al.(2024)Plaat, Wong, Verberne, Broekens, van Stein, and Back}]{plaat2024reasoning}
Aske Plaat, Annie Wong, Suzan Verberne, Joost Broekens, Niki van Stein, and Thomas Back. 2024.
\newblock Reasoning with large language models, a survey.
\newblock \emph{arXiv preprint arXiv:2407.11511}.

\bibitem[{Quan et~al.(2025)Quan, Yang, Yu, Zheng, Liu, Yang, Ren, Gao, Miao, Feng et~al.}]{quan2025codeelo}
Shanghaoran Quan, Jiaxi Yang, Bowen Yu, Bo~Zheng, Dayiheng Liu, An~Yang, Xuancheng Ren, Bofei Gao, Yibo Miao, Yunlong Feng, et~al. 2025.
\newblock Codeelo: Benchmarking competition-level code generation of llms with human-comparable elo ratings.
\newblock \emph{arXiv preprint arXiv:2501.01257}.

\bibitem[{Sanh et~al.(2022)Sanh, Webson, Raffel, Bach, Sutawika, Alyafeai, Chaffin, Stiegler, Raja, Dey et~al.}]{sanhmultitask}
Victor Sanh, Albert Webson, Colin Raffel, Stephen Bach, Lintang Sutawika, Zaid Alyafeai, Antoine Chaffin, Arnaud Stiegler, Arun Raja, Manan Dey, et~al. 2022.
\newblock Multitask prompted training enables zero-shot task generalization.
\newblock In \emph{International Conference on Learning Representations}.

\bibitem[{Team et~al.(2024)Team, Mesnard, Hardin, Dadashi, Bhupatiraju, Pathak, Sifre, Rivi{\`e}re, Kale, Love et~al.}]{team2024gemma}
Gemma Team, Thomas Mesnard, Cassidy Hardin, Robert Dadashi, Surya Bhupatiraju, Shreya Pathak, Laurent Sifre, Morgane Rivi{\`e}re, Mihir~Sanjay Kale, Juliette Love, et~al. 2024.
\newblock Gemma: Open models based on gemini research and technology.
\newblock \emph{arXiv preprint arXiv:2403.08295}.

\bibitem[{Team(2024)}]{Open-O1}
Open-O1 Team. 2024.
\newblock Open-o1: A model for matching proprietary openai o1's power with open-source innovation.
\newblock Accessed: 2025-02-08.

\bibitem[{Wang et~al.(2022)Wang, Mishra, Alipoormolabashi, Kordi, Mirzaei, Naik, Ashok, Dhanasekaran, Arunkumar, Stap et~al.}]{wang2022super}
Yizhong Wang, Swaroop Mishra, Pegah Alipoormolabashi, Yeganeh Kordi, Amirreza Mirzaei, Atharva Naik, Arjun Ashok, Arut~Selvan Dhanasekaran, Anjana Arunkumar, David Stap, et~al. 2022.
\newblock Super-naturalinstructions: Generalization via declarative instructions on 1600+ nlp tasks.
\newblock In \emph{Proceedings of the 2022 Conference on Empirical Methods in Natural Language Processing}, pages 5085--5109.

\bibitem[{Wei et~al.(2022{\natexlab{a}})Wei, Bosma, Zhao, Guu, Yu, Lester, Du, Dai, and Le}]{weifinetuned}
Jason Wei, Maarten Bosma, Vincent Zhao, Kelvin Guu, Adams~Wei Yu, Brian Lester, Nan Du, Andrew~M Dai, and Quoc~V Le. 2022{\natexlab{a}}.
\newblock Finetuned language models are zero-shot learners.
\newblock In \emph{International Conference on Learning Representations}.

\bibitem[{Wei et~al.(2022{\natexlab{b}})Wei, Wang, Schuurmans, Bosma, Xia, Chi, Le, Zhou et~al.}]{wei2022chain}
Jason Wei, Xuezhi Wang, Dale Schuurmans, Maarten Bosma, Fei Xia, Ed~Chi, Quoc~V Le, Denny Zhou, et~al. 2022{\natexlab{b}}.
\newblock Chain-of-thought prompting elicits reasoning in large language models.
\newblock \emph{Advances in neural information processing systems}, 35:24824--24837.

\bibitem[{Wu et~al.(2024)Wu, Peng, Du, Zheng, Liu, Wu, Ma, Li, Yang, Zhou et~al.}]{wu2024comparative}
Siwei Wu, Zhongyuan Peng, Xinrun Du, Tuney Zheng, Minghao Liu, Jialong Wu, Jiachen Ma, Yizhi Li, Jian Yang, Wangchunshu Zhou, et~al. 2024.
\newblock A comparative study on reasoning patterns of openai's o1 model.
\newblock \emph{arXiv preprint arXiv:2410.13639}.

\bibitem[{Xiong et~al.(2023)Xiong, Ding, Cao, Liu, and Qin}]{xiong2023examining}
Kai Xiong, Xiao Ding, Yixin Cao, Ting Liu, and Bing Qin. 2023.
\newblock Examining inter-consistency of large language models collaboration: An in-depth analysis via debate.
\newblock In \emph{Findings of the Association for Computational Linguistics: EMNLP 2023}, pages 7572--7590.

\bibitem[{Xiong et~al.(2024{\natexlab{a}})Xiong, Ding, Du, Ying, Liu, Qin, and Cao}]{xiong2024diagnosing}
Kai Xiong, Xiao Ding, Li~Du, Jiahao Ying, Ting Liu, Bing Qin, and Yixin Cao. 2024{\natexlab{a}}.
\newblock Diagnosing and remedying knowledge deficiencies in llms via label-free curricular meaningful learning.
\newblock \emph{arXiv preprint arXiv:2408.11431}.

\bibitem[{Xiong et~al.(2022)Xiong, Ding, Li, Du, Liu, Qin, Zheng, and Huai}]{xiong2022reco}
Kai Xiong, Xiao Ding, Zhongyang Li, Li~Du, Ting Liu, Bing Qin, Yi~Zheng, and Baoxing Huai. 2022.
\newblock Reco: Reliable causal chain reasoning via structural causal recurrent neural networks.
\newblock In \emph{Proceedings of the 2022 Conference on Empirical Methods in Natural Language Processing}, pages 6426--6438.

\bibitem[{Xiong et~al.(2024{\natexlab{b}})Xiong, Ding, Liu, Qin, Xu, Yang, Liu, and Cao}]{xiong2024meaningful}
Kai Xiong, Xiao Ding, Ting Liu, Bing Qin, Dongliang Xu, Qing Yang, Hongtao Liu, and Yixin Cao. 2024{\natexlab{b}}.
\newblock Meaningful learning: Enhancing abstract reasoning in large language models via generic fact guidance.
\newblock In \emph{Advances in Neural Information Processing Systems}, volume~37, pages 120501--120525.

\bibitem[{Xu et~al.(2024)Xu, Sun, Zheng, Geng, Zhao, Feng, Tao, Lin, and Jiang}]{xu2024wizardlm}
Can Xu, Qingfeng Sun, Kai Zheng, Xiubo Geng, Pu~Zhao, Jiazhan Feng, Chongyang Tao, Qingwei Lin, and Daxin Jiang. 2024.
\newblock Wizardlm: Empowering large pre-trained language models to follow complex instructions.
\newblock In \emph{The Twelfth International Conference on Learning Representations}.

\bibitem[{Yang et~al.(2024)Yang, Yang, Zhang, Hui, Zheng, Yu, Li, Liu, Huang, Wei et~al.}]{yang2024qwen2}
An~Yang, Baosong Yang, Beichen Zhang, Binyuan Hui, Bo~Zheng, Bowen Yu, Chengyuan Li, Dayiheng Liu, Fei Huang, Haoran Wei, et~al. 2024.
\newblock Qwen2. 5 technical report.
\newblock \emph{arXiv preprint arXiv:2412.15115}.

\bibitem[{Yao et~al.(2024)Yao, Yu, Zhao, Shafran, Griffiths, Cao, and Narasimhan}]{yao2024tree}
Shunyu Yao, Dian Yu, Jeffrey Zhao, Izhak Shafran, Tom Griffiths, Yuan Cao, and Karthik Narasimhan. 2024.
\newblock Tree of thoughts: Deliberate problem solving with large language models.
\newblock \emph{Advances in Neural Information Processing Systems}, 36.

\bibitem[{Yao et~al.(2023)Yao, Zhao, Yu, Du, Shafran, Narasimhan, and Cao}]{yaoreact}
Shunyu Yao, Jeffrey Zhao, Dian Yu, Nan Du, Izhak Shafran, Karthik~R Narasimhan, and Yuan Cao. 2023.
\newblock React: Synergizing reasoning and acting in language models.
\newblock In \emph{The Eleventh International Conference on Learning Representations}.

\bibitem[{Ying et~al.(2024)Ying, Cao, Xiong, Cui, He, and Liu}]{ying2024intuitive}
Jiahao Ying, Yixin Cao, Kai Xiong, Long Cui, Yidong He, and Yongbin Liu. 2024.
\newblock Intuitive or dependent? investigating llms’ behavior style to conflicting prompts.
\newblock In \emph{Proceedings of the 62nd Annual Meeting of the Association for Computational Linguistics (Volume 1: Long Papers)}, pages 4221--4246.

\bibitem[{Zhang et~al.(2023)Zhang, Zhang, Li, and Smola}]{zhangautomatic}
Zhuosheng Zhang, Aston Zhang, Mu~Li, and Alex Smola. 2023.
\newblock Automatic chain of thought prompting in large language models.
\newblock In \emph{The Eleventh International Conference on Learning Representations}.

\bibitem[{Zhao et~al.(2024)Zhao, Yin, Zeng, Wang, Shi, Lyu, Wang, Luo, and Zhang}]{zhao2024marco}
Yu~Zhao, Huifeng Yin, Bo~Zeng, Hao Wang, Tianqi Shi, Chenyang Lyu, Longyue Wang, Weihua Luo, and Kaifu Zhang. 2024.
\newblock Marco-o1: Towards open reasoning models for open-ended solutions.
\newblock \emph{arXiv preprint arXiv:2411.14405}.

\bibitem[{Zhou et~al.(2023)Zhou, Sch{\"a}rli, Hou, Wei, Scales, Wang, Schuurmans, Cui, Bousquet, Le et~al.}]{zhouleast}
Denny Zhou, Nathanael Sch{\"a}rli, Le~Hou, Jason Wei, Nathan Scales, Xuezhi Wang, Dale Schuurmans, Claire Cui, Olivier Bousquet, Quoc~V Le, et~al. 2023.
\newblock Least-to-most prompting enables complex reasoning in large language models.
\newblock In \emph{The Eleventh International Conference on Learning Representations}.

\end{thebibliography}

\appendix
\newpage
\onecolumn
\section{Prompts for Com$^2$ Creation}
\label{app:com2_prompt}

\begin{tcolorbox}[title={Event Proposal},fontupper = \scriptsize]
    You are given some example events, you should follow the pattern of the example events to create lots of new events.
    Example Events:\\
    \{\}\\
    You can create new events in any areas (such as science, math, economy, health, daily, sports, art, etc.) and scenarios. Just give the created new events and create as many events as possible. You format should be like:\\
    1.\\
    2.\\
    3.\\
    ······
\end{tcolorbox}

\begin{tcolorbox}[title={Causal Chain Proposal: Direct},fontupper = \scriptsize]
You are an expert in causality. Please create a causal chains based on the given event. The following are the requirements:\\
1. the causal chain contains 5 events and 4 causal relationships.\\
2. the causal chain should be common in real life.\\
3. the causal chain should has a proper time gap to make users cannot infer tail event just based on the given event.\\
Here are the given event:\\
\{\}\\
You should follow the format like:\\
\{\} -> EVENT -> EVENT -> EVENT -> EVENT
\end{tcolorbox}

\begin{tcolorbox}[title={Causal Chain Proposal: Intervention},fontupper = \scriptsize]
You are an expert in causality. Please create two causal chains based on and start from the given event. The following are the requirements:\\
1. each causal chain contains 5 events and 4 causal relationships, each chain should be reasonable.\\
2. the first causal chain should have a high probability to happen, which means it is common in real life.\\
3. the second causal chain should have a lower probality to happend than the first chain.\\
4. each causal chain should has a proper time gap to make users cannot infer tail event just based on the given event.\\
5. the tail event should be the same.\\

Here are the given event:\\
\{\}\\
You should follow the format like:\\
High Probability: {} -> EVENT1 -> EVENT2 -> EVENT3 -> SAME EVENT\\
Low Probability: {} -> EVENT4 -> EVENT5 -> EVENT6 -> SAME EVENT
\end{tcolorbox}

\begin{tcolorbox}[title={Causal Chain Proposal: Counterfactual},fontupper = \scriptsize]
You are an expert in causality. Please create two causal chains based on the given event. The following are the requirements:\\
1. each causal chain contains 5 events and 4 causal relationships.\\
2. the first causal chain should be common in real life.\\
3. the second causal chain should be a counterfactual scenario (science fiction or an event which is unable to intervene) of the first causal chain.\\
4. each causal chain should has a proper time gap to make users cannot infer tail event just based on the given event.\\

Here are the given event:\\
\{\}\\
You should follow the format like:\\
Normal: \{\} -> EVENT -> EVENT -> EVENT -> EVENT\\
Counterfactual: Counterfactual\_of\_given\_event -> EVENT -> EVENT -> EVENT -> EVENT
\end{tcolorbox}

\begin{tcolorbox}[title={Causal Chain Proposal: Decision},fontupper = \scriptsize]
You are an expert in causality. Please create two causal chains based on and start from the given event. The following are the requirements:\\
1. each causal chain contains 5 events and 4 causal relationships, each chain should be reasonable.\\
2. each chain would lead to bad outcomes or something unacceptable.\\
3. the first causal chain should have a high probability to happen, which means it is common in real life.\\
4. the second causal chain should have a much lower probality to happend than the first chain.\\
5. each causal chain should has a proper time gap to make users cannot infer tail event just based on the given event.\\

Here are the given event:\\
\{\}
You should follow the format like:\\
High Probability: \{\} -> EVENT -> EVENT -> EVENT -> EVENT\\
Low Probability: \{\} -> EVENT -> EVENT -> EVENT -> EVENT
\end{tcolorbox}

\begin{tcolorbox}[title={Causal Chain Proposal: Transition},fontupper = \scriptsize]
You are an expert in causality. Please create three causal chains based on and start from the given event. The following are the requirements:\\
1. each causal chain contains 5 events and 4 causal relationships.\\
2. the first causal chain should be reasonable without any causal transitive problems.\\
3. the second causal chain should contain a scene drift problem.\\
4. the thrid causal chain should contain a threshold effect problem.\\
5. each causal chain should has a proper time gap to make users cannot infer tail event just based on the given event.\\

Take a causal chian A -> B -> C for example, here are the reference of the reason for the causal transitive problem:\\
1. scene drift: A -> B and B -> C would not happen within the same specific scene.\\
2. threshold effect: the influence of A on B is not enough for B to cause C.\\

Here are the given event:\\
\{\}\\
Just give the chains only. You should follow the format like:\\
Normal: {} -> EVENT -> EVENT -> EVENT -> EVENT\\
With Scene Drift: {} -> EVENT -> EVENT -> EVENT -> EVENT\\
With Threshold Effect: {} -> EVENT -> EVENT -> EVENT -> EVENT
\end{tcolorbox}

\begin{tcolorbox}[title={Causal Graph Proposal and Com$^2$ Synthesis: Direct},fontupper = \small]
After understanding the casual chain, we need you to create a multiple-choice example based on the causal chain with a slow thinking process. The following are the requirements:\\
1. the example should contain a question, a slow thinking process, some options, and an answer.\\
2. note that the causal chain is not a part of the final example, it is just used to help you design the example.\\
3. in the question, you should create a suitable question. The intermediate events in the causal chains cannot appear in the question. The question is asking for a most plausible outcome.\\
4. the final answer should be the tail event of the causal chain, and the other wrong options should be deceptive. \\
5. based on the causal chain, you should create a slowing thinking process, which consists of several actions (selecting from the following actions), actions can be repeated:
    \begin{adjustwidth}{0.6cm}{0cm}
        -Systematic Analysis: Starting from the overall structure of the problem, first analyze the inputs and outputs, as well as the constraints, and then decide on the choice of algorithm and the use of data structures.\\
        -Method Reuse: For some problems that can be transformed into classic problems (such as the shortest path or knapsack problem), you can quickly reuse existing methods to solve them.\\
        -Comparison: If the question is a multiple-choice question, you should compare the differences among all options in detail based on the question.\\
        -Divide: Break down a complex causal-related problem into subproblems.\\
        -Conquer: Constructs the overall solution by repeatedly solving each subproblem step-by-step with the guidance of causal chain (note that it will not given in the question).\\
        -Self-Refinement: Assessing your reasoning process during inference to determine if there are any issues and correct any errors. Using "wait" to start this.\\
        -Context Identification: For some datasets requiring additional information input, you first summarizes different aspects of the context related to the question, and then gives the response for the corresponding question.\\
        -Emphasizing Constraints: For some datasets with constraints on the generated text, emphasizes the corresponding constraints during the reasoning process.
    \end{adjustwidth}
6. the whole slow thinking process should simulate first-person thinking.\\
7. do not explicitly demonstrate which chain you are using, all the chains are used to guide the slowing thinking process, it will not be given to help with question answering. \\
8. the created question should be as concise as possible, and the slow thinking process should be as detailed and complex as possible.\\
9. you should make a **mistake** first, and conduct self-refinement to backtrack and reason correctly by the causal chain.\\
10. the output format should be: Question: \_\_\textbackslash n\textbackslash n Options: \_\_\textbackslash n\textbackslash n Slow Thinking Process: \_\_\textbackslash n\textbackslash n Answer: \_\_.
\end{tcolorbox}

\begin{tcolorbox}[title={Causal Graph Proposal and Com$^2$ Synthesis: Intervention},fontupper = \small]
After understanding the causal chains, we need you to create a multiple-choice example. The following are the requirements:\\ 
1. the example should contain a question, a slow thinking process, some options, and an answer.\\
2. note that the causal chains are not a part of the final example, they are just used to help you design the example.\\
3. the intermediate events in each causal chain cannot appear in the question.\\
4. you should conduct an intervention on the first causal chain to achieve the following goals:
    \begin{adjustwidth}{0.6cm}{0cm}
        -in the first causal chain, the transit to the tail event is interrupted by the intervention.\\
        -in the second causal chain, the transit is NOT interrupted by the intervention.
    \end{adjustwidth}
4. the question can only contain the first events of the causal chains and the intervention event.\\
5. the final answer should be the tail event of the second causal chain, and the other options should be wrong. \\
6. based on the causal chains, you should create a slowing thinking process, which consists of several actions (selecting from the following actions), actions can be repeated:
    \begin{adjustwidth}{0.6cm}{0cm}
    -Systematic Analysis: Starting from the overall structure of the problem, first analyze the inputs and outputs, as well as the constraints, and then decide on the choice of algorithm and the use of data structures.\\
    -Method Reuse: For some problems that can be transformed into classic problems (such as the shortest path or knapsack problem), you can quickly reuse existing methods to solve them.\\
    -Comparison: If the question is a multiple-choice question, you should compare the differences among all options in detail based on the question.\\
    -Divide: Break down a complex causal-related problem into subproblems.\\
    -Conquer: Constructs the overall solution by repeatedly solving each subproblem step-by-step with the guidance of causal chain (note that it will not given in the question).\\
    -Self-Refinement: Assessing your reasoning process during inference to determine if there are any issues and correct any errors. For example, conduct reflection on the intervened position and continue the reasoning via another chain.\\
    -Context Identification: For some datasets requiring additional information input, you first summarizes different aspects of the context related to the question, and then gives the response for the corresponding question.\\
    -Emphasizing Constraints: For some datasets with constraints on the generated text, emphasizes the corresponding constraints during the reasoning process.
    \end{adjustwidth}
7. the whole slow thinking process should simulate first-person thinking.\\
8. the created question should be as concise as possible, and the slow thinking process should be as detailed and complex as possible.\\
9. the output format should be: Question: \_\_\textbackslash n\textbackslash n Options: \_\_\textbackslash n\textbackslash n Slow Thinking Process: \_\_\textbackslash n\textbackslash n Answer: \_\_.
\end{tcolorbox}

\begin{tcolorbox}[title={Causal Graph Proposal and Com$^2$ Synthesis: Counterfactual},fontupper = \scriptsize]
After understanding the causal chains, we need you to create a multiple-choice example. The following are the requirements:\\
1. the example should contain a question, a slow thinking process, some options, and an answer.\\
2. in the question, the intermediate events in each causal chain cannot appear in the question.\\
3. note that the causal chains are not a part of the final example, they are just used to help you design the example.\\
4. you should use the counterfactual chain to create a counterfactual question (e.g. If A leads to B not C, what will happen?).\\
5. the final answer should be the tail event of the counterfactual chain, and the other wrong options should be deceptive.\\
6. based on the causal chains, you should simulate first-person thinking and create a slowing thinking process, which consists of several actions (selecting from the following actions), actions can be repeated:
\begin{adjustwidth}{0.6cm}{0cm}
    -Systematic Analysis: Starting from the overall structure of the problem, first analyze the inputs and outputs, as well as the constraints, and then decide on the choice of algorithm and the use of data structures.\\
    -Method Reuse: For some problems that can be transformed into classic problems (such as the shortest path or knapsack problem), you can quickly reuse existing methods to solve them.\\
    -Comparison: If the question is a multiple-choice question, you should compare the differences among all options in detail based on the question. Moreover, you should compare the real condition to the counterfactual to obtain detailed analysis.\\
    -Divide: Break down a complex causal-related problem into subproblems.\\
    -Conquer: Constructs the overall solution by repeatedly solving each subproblem step-by-step with the guidance of causal chain (note that it will not given in the question).\\
    -Self-Refinement: Assessing your reasoning process during inference to determine if there are any issues and correct any errors. Using "wait" to start this.\\
    -Context Identification: For some datasets requiring additional information input, you first summarizes different aspects of the context related to the question, and then gives the response for the corresponding question.\\
    -Emphasizing Constraints: For some datasets with constraints on the generated text, emphasizes the corresponding constraints during the reasoning process.
\end{adjustwidth}
7. the created question should be as concise as possible, and the slow thinking process should be as detailed and complex as possible.\\
8. the counterfactual must be conducted in the question, which cannot appear in the thinking process.\\
9. you should make a **mistake** first, and conduct self-refinement to backtrack and reason correctly.\\
10. the output format should be: Question: \_\_\textbackslash n\textbackslash n Options: \_\_\textbackslash n\textbackslash n Slow Thinking Process: \_\_\textbackslash n\textbackslash n Answer: \_\_.
\end{tcolorbox}

\begin{tcolorbox}[title={Causal Graph Proposal and Com$^2$ Synthesis: Decision},fontupper = \scriptsize]
After understanding the causal chains carefully, we need you to create a multiple-select example. The following are the requirements:\\
1. the example should contain a question, a slow thinking process, several options, and answers.\\
2. note that the causal chains are not a part of the final example, they are just used to help you design the example.\\
3. the question mainly investigates how to severe results, which means what interventions we can do to break the transition to bad results.\\
4. the answer choices should be interventions applied to the intermediate events of the given causal chains.\\
5. in the options, besides the answers choices, you should provide several distractors, which are wrong but deceptive.\\
6. the intermediate events in the causal chains cannot appear in the question.\\
7. based on the causal chains and question, you should create a slowing thinking process, which consists of several actions (selecting from the following actions), actions can be repeated:
\begin{adjustwidth}{0.6cm}{0cm}
    -Systematic Analysis: Starting from the overall structure of the problem, first analyze the inputs and outputs, as well as the constraints, and then decide on the choice of algorithm and the use of data structures.\\
    -Method Reuse: For some problems that can be transformed into classic problems (such as the shortest path or knapsack problem), you can quickly reuse existing methods to solve them.\\
    -Comparison: If the question is a multiple-select question, you should analysis each option in detail based on the question.\\
    -Divide: Break down a complex causal-related problem into subproblems.\\
    -Conquer: Constructs the overall solution by repeatedly solving each subproblem step-by-step with the guidance of causal chain (note that it will not given in the question).\\
    -Self-Refinement: Assessing your reasoning process during inference to determine if there are any issues and correct any errors.\\
    -Context Identification: For some datasets requiring additional information input, you first summarizes different aspects of the context related to the question, and then gives the response for the corresponding question.\\
    -Emphasizing Constraints: For some datasets with constraints on the generated text, emphasizes the corresponding constraints during the reasoning process.
\end{adjustwidth}
8. the whole slow thinking process should simulate first-person thinking.\\
9. the created question should be as concise as possible, while the slow thinking process should be as detailed and complex as possible.\\
10. the output format should be: Question: \_\_\textbackslash n\textbackslash n Options: \_\_\textbackslash n\textbackslash n Slow Thinking Process: \_\_\textbackslash n\textbackslash n Answer: \_\_.
\end{tcolorbox}

\begin{tcolorbox}[title={Causal Graph Proposal and Com$^2$ Synthesis: Transition},fontupper = \scriptsize]
After understanding the causal chains and the scenario, we need you to create a multiple-choice example. The following are the requirements: \\
1. the example should contain a question, a slow thinking process, some options, and an answer.\\
2. note that the causal chains and scenario are not a part of the final example, they are just used to help you design the example.\\
3. the intermediate events in each causal chain cannot appear in the question.\\
4. the second has the causal transitive problem of {}, take A -> B -> C for example, here are the definitions:
\begin{adjustwidth}{0.6cm}{0cm}
    -scene drift: A -> B and B -> C would not happen within the same specific scene.\\
    -threshold effect: the influence of A on B is not enough for B to cause C.
\end{adjustwidth}
4. the question can only contain the first events of the causal chains.\\
5. the final answer should be the tail event of the first causal chain, and the other options should be wrong.\\
6. based on the causal chains and scenario, you should create a slowing thinking process, which consists of several actions (selecting from the following actions), actions can be repeated:
\begin{adjustwidth}{0.6cm}{0cm}
    -Systematic Analysis: Starting from the overall structure of the problem, first analyze the inputs and outputs, as well as the constraints, and then decide on the choice of algorithm and the use of data structures.\\
    -Method Reuse: For some problems that can be transformed into classic problems (such as the shortest path or knapsack problem), you can quickly reuse existing methods to solve them.\\
    -Comparison: If the question is a multiple-choice question, you should compare the differences among all options in detail based on the question.\\
    -Divide: Break down a complex causal-related problem into subproblems.\\
    -Conquer: Constructs the overall solution by repeatedly solving each subproblem step-by-step with the guidance of causal chain (note that it will not given in the question).\\
    -Self-Refinement: Assessing your reasoning process during inference to determine if there are any issues and correct any errors. Using "wait" to start this.\\
    -Context Identification: For some datasets requiring additional information input, you first summarizes different aspects of the context related to the question, and then gives the response for the corresponding question.\\
    -Emphasizing Constraints: For some datasets with constraints on the generated text, emphasizes the corresponding constraints during the reasoning process.
\end{adjustwidth}
7. you should conduct reasoning via the second causal chain to make a **mistake** and conduct self-refinement to backtrack and reason by the second causal chain.\\
8. the whole slow thinking process should simulate first-person thinking.\\
9. the created question should be as concise as possible, and the slow thinking process should be as detailed and complex as possible.\\
10. the output format should be: Question: \_\_\textbackslash n\textbackslash n Options: \_\_\textbackslash n\textbackslash n Slow Thinking Process: \_\_\textbackslash n\textbackslash n Answer: \_\_.
\end{tcolorbox}

\begin{tcolorbox}[title={Detective Proposal and Com$^2$ Synthesis: Intervention}, fontupper = \scriptsize]
    \textbf{criminal} \\
    \{\}\textbackslash n-------\textbackslash nWho is(are) the criminal(s)? Just output the name(s). \\
    \\
    \textbf{puzzle} \\
    This is a detective story, can you generate a puzzle based on the story, which contains a statement (a short paragraph) of the crime, all the initial facts and people's actions found by police and detectives (not inferred clues, and should exclude \{\}), which means the police and detectives can infer the criminals just based on the facts and actions rather than the story. Please give a puzzle to ask who is (are) the criminal(s). Some suspects. Please follow the format like: Crime statement: \_\_\textbackslash n Facts and actions: \_\_\textbackslash n Puzzle: \_\_\textbackslash n Suspects: (indexed by A), B), C) and so on, list all suspects, should include \{\}. Do not provide answer. The last option in the suspects should be "None of the above". \\
    \\
    \textbf{Slow Thinking} \\
    Please follow the facts and actions only to give a step-by-step investigation to describe a not guilty person as the criminal (means you should make a mistake first). And then you are aware of an essential clue that "\{\}", you start to reflect your mistakes, finally, you got the right answer by backtracking and reasoning. The response should end with "Answer: (the index + the option content, such as A) Tom)". The answer MUST be "\{\}".
\end{tcolorbox}

\begin{tcolorbox}[title={Detective Proposal and Com$^2$ Synthesis: Counterfactual}, fontupper = \scriptsize]
    \textbf{criminal} \\
    \{\}\textbackslash n-------\textbackslash nWho is(are) the criminal(s)? Just output the name(s).
    \\
    \textbf{puzzle} \\
    This is a detective story, please generate a question based on the story, which contains a statement (a short paragraph) of the crime, all the initial facts and people's actions found by police and detectives (not inferred clues), which means the police and detectives can infer the criminals just based on the facts and actions rather than the story. Please give a multi-choice **counterfactual** question, A slowing thinking process which consists of several actions (selecting from the following actions), actions can be repeated, the whole process should use detailed information in the statement, facts and actions for detailed analysis:
    \begin{adjustwidth}{0.6cm}{0cm}
        -Systematic Analysis: Starting from the overall structure of the problem, first analyze the inputs and outputs, as well as the constraints, and then decide on the choice of algorithm and the use of data structures.\\
        -Method Reuse: For some problems that can be transformed into classic problems (such as the shortest path or knapsack problem), you can quickly reuse existing methods to solve them.\\
        -Comparison: If the question is a multiple-select question, you should analysis each option in detail based on the question.\\
        -Divide: Break down a complex causal-related problem into subproblems.\\
        -Conquer: Constructs the overall solution by repeatedly solving each subproblem step-by-step with the guidance of causal chain (note that it will not given in the question).\\
        -Self-Refinement: Assessing your reasoning process during inference to determine if there are any issues and correct any errors.\\
        -Context Identification: For some datasets requiring additional information input, you first summarizes different aspects of the context related to the question, and then gives the response for the corresponding question.\\
        -Emphasizing Constraints: For some datasets with constraints on the generated text, emphasizes the corresponding constraints during the reasoning process.
    \end{adjustwidth}
    Please follow the format like: Crime statement: \_\_\textbackslash nFacts and actions: \_\_\textbackslash nQuestion: \_\_\textbackslash nOptions: (indexed by A), B),C) and so on, include some wrong options)\textbackslash nSlow Thinking Process: \_\_\textbackslash nAnswer:\_\_(index + option text).
\end{tcolorbox}

\begin{tcolorbox}[title={Detective Proposal and Com$^2$ Synthesis: Decision}, fontupper = \scriptsize]
    \textbf{criminal} \\
    \{\}\textbackslash n-------\textbackslash nWho is(are) the criminal(s)? Just output the name(s).
    \\
    \textbf{puzzle} \\
    This is a detective story, please generate a question based on the story, which contains a statement (a short paragraph) of the crime, all the initial facts and people's actions found by police and detectives (not inferred clues), which means the police and detectives can infer the criminals just based on the facts and actions rather than the story. Please give a multi-select question to ask what we can have done to prevent this crime from happening. A slowing thinking process which consists of several actions (selecting from the following actions), actions can be repeated, the whole process should use detailed information in the statement, facts and actions for detailed analysis:
    \begin{adjustwidth}{0.6cm}{0cm}
        -Systematic Analysis: Starting from the overall structure of the problem, first analyze the inputs and outputs, as well as the constraints, and then decide on the choice of algorithm and the use of data structures.\\
        -Method Reuse: For some problems that can be transformed into classic problems (such as the shortest path or knapsack problem), you can quickly reuse existing methods to solve them.\\
        -Comparison: If the question is a multiple-select question, you should analysis each option in detail based on the question.\\
        -Divide: Break down a complex causal-related problem into subproblems.\\
        -Conquer: Constructs the overall solution by repeatedly solving each subproblem step-by-step with the guidance of causal chain (note that it will not given in the question).\\
        -Self-Refinement: Assessing your reasoning process during inference to determine if there are any issues and correct any errors.\\
        -Context Identification: For some datasets requiring additional information input, you first summarizes different aspects of the context related to the question, and then gives the response for the corresponding question.\\
        -Emphasizing Constraints: For some datasets with constraints on the generated text, emphasizes the corresponding constraints during the reasoning process.
    \end{adjustwidth}
    Please follow the format like: Crime statement: \_\_\textbackslash nFacts and actions: \_\_\textbackslash nQuestion: \_\_\textbackslash nOptions: (indexed by A), B),C) and so on, include some wrong options)\textbackslash nSlow Thinking Process: \_\_\textbackslash nAnswer:\_\_(index + option text).
\end{tcolorbox}
\section{Details of Human Evaluation}
\label{app:human_eval}
We choose three annotators with good backgrounds in causal inference and commonsense reasoning. We randomly and respectively sample 100 examples of various tasks in Com$^2$-main and Com$^2$-hard for the human evaluation. Thus, we evaluate the slow thinking on these 200 examples. For each annotator, we pay \$10 per hour, while in our country, the minimum wage is less than \$5 per hour. All annotators agreed to let us use their annotations. The evaluation agreements are 94\%, 91\%, and 90\% on the evaluations of Com$^2$-main, Com$^2$-hard, and slow thinking, respectively.
\section{Examples in Com$^2$}
\label{app:examples}

\begin{tcolorbox}[title={Direct in Com$^2$-main},fontupper = \scriptsize]
\textbf{Question}\\
If Sarah had ignored her symptoms completely, what would eventually happen?\\
\textbf{Options}\\
A) She would feel better after a short period.  \textbackslash nB) She would schedule a follow-up appointment.  \textbackslash nC) She would experience a sudden health crisis.  \textbackslash nD) She would receive a prescription for pain relief.\\
\textbf{Correct Answer}\\
C) She would experience a sudden health crisis.\\
\\
\textbf{Slow Thinking}\\
1. Context Identification: I need to analyze the situation where Sarah ignored her symptoms. Ignoring health issues typically leads to worsening conditions. I need to determine the long-term consequences of ignoring symptoms rather than seeking help. \\
2. Systematic Analysis: Let’s break down the potential outcomes of ignoring symptoms. One possibility is that she could start feeling better due to the body's natural healing process. Another option could be that she decides to just cope with the pain without any medical intervention.\\
3. Comparison: I’ll compare all the options. Option A suggests she would feel better, which often isn’t realistic with persistent symptoms. Option B indicates she would schedule a follow-up appointment, but if she’s ignoring the symptoms, this doesn’t align with her behavior. Option D implies she might seek relief through medication, which contradicts the act of ignoring symptoms entirely.\\
4. Divide: Breaking down the outcomes, I realize that not addressing health issues often leads to escalation. If she continues to ignore symptoms, it’s likely her situation will worsen instead of improving.\\
5. Conquer: I need to construct a logical conclusion about the most serious possibility. By ignoring the symptoms, one possible severe consequence is a sudden health crisis necessitating emergency intervention.\\
6. Self-Refinement: Wait, I almost concluded that she might just feel better (Option A), but that’s an optimistic assumption. Considering realistic scenarios, ignoring serious symptoms often leads to drastic health declines rather than recovery. I need to reassess.\\
7. Final Assessment: Conclusively, experiencing a sudden health crisis is the most likely negative outcome of ignoring persistent health issues.
\end{tcolorbox}

\begin{tcolorbox}[title={Decison in Com$^2$-main},fontupper = \scriptsize]
\textbf{Question}\\
What interventions can prevent severe consequences during a hot air ballooning experience that could lead to disastrous outcomes?\\
\textbf{Options}\\
A. Regular equipment checks before the flight  \textbackslash nB. Ignoring weather warnings and proceeding with the flight  \textbackslash nC. Ensuring pilots undergo thorough training for extreme conditions  \textbackslash nD. Allowing passengers to control the balloon  \textbackslash nE. Establishing no-fly zones in higher-density airspace  \textbackslash nF. Offering balloon rides regardless of adverse weather conditions
\textbf{Correct Answer}\\
A, C, E.\\
\\
\textbf{Slow Thinking}\\
1. Context Identification: I first recognize that the question revolves around potential interventions that could mitigate severe negative outcomes during a hot air ballooning experience. I need to focus on preventing issues related to equipment failure and extreme weather.\\
2. Systematic Analysis: I review each option in light of their potential to prevent accidents. Option A refers to equipment checks and could help avert malfunctions, while Option C emphasizes pilot training during severe conditions.\\
3. Comparison: I analyze options B and F. Both suggest ignoring critical safety precautions and could lead to disastrous outcomes. Option D seems reckless, as it suggests letting passengers take control without proper training.\\
4. Divide: Next, I break down the options into two groups: safety interventions and unsafe practices. Grouping this way highlights which options can truly help prevent severe consequences.\\
5. Conquer: I take the valid interventions from the safe group (A, C, E) and recognize their importance in preventing negative outcomes associated with the hazards of hot air ballooning.\\
6. Self-Refinement: I reassess my thought process, making certain that I haven’t overlooked any relevant options or context, confirming that my focus remains on preventing equipment failure and addressing pilot preparedness.\\
7. Emphasizing Constraints: I remember that the question explicitly seeks interventions that will reduce severe outcomes, which validates the importance of option E about establishing no-fly zones.\\
8. Method Reuse: I recall existing safety protocols in aviation, reaffirming that rigorous checks and pilot training are part of best practices in preventing accidents.
\end{tcolorbox}

\begin{tcolorbox}[title={Transition in Com$^2$-main},fontupper = \scriptsize]
\textbf{Question}\\
What is the first step to build a bookshelf from scratch?\\
\textbf{Options}\\
A) Paint or stain the finished bookshelf  \textbackslash nB) Gather materials and tools  \textbackslash nC) Attend a woodworking workshop  \textbackslash nD) Share experiences on social media\\
\textbf{Correct Answer}\\
B) Gather materials and tools.\\
\\
\textbf{Slow Thinking}\\
I start by identifying the question, which asks for the first step in building a bookshelf from scratch. This is a straightforward task, but I need to ensure I analyze the options carefully.  \\
- Comparison: I look at the options provided. Option A mentions painting or staining, which clearly cannot be the first step since it comes after the bookshelf is constructed. Option C involves attending a workshop, which is not directly related to the act of building the bookshelf itself. Option D talks about sharing experiences, which is also a post-construction activity.  \\
- Divide: I break down the remaining options. Option B, \"Gather materials and tools,\" seems to be the only logical first step.  \\
- Context Identification: I think about the context of building a bookshelf. The process typically starts with gathering the necessary items before any construction can begin.  \\
- Self-Refinement: I pause to assess my reasoning. I realize that I might have overlooked the possibility of needing to plan the design first, but that would still require materials and tools to execute the plan.  \\
- Conquer: I conclude that gathering materials and tools is indeed the first action needed to start the project.  \\
After this thorough analysis, I am confident that the correct answer is B.
\end{tcolorbox}

\begin{tcolorbox}[title={Intervention in Com$^2$-main},fontupper = \scriptsize]
\textbf{Question}\\
A child snaps a twig off a branch, and a nearby animal hears the snap and perceives danger. What happens next?\\
\textbf{Options}\\
A) The bird returns to its nest with food for its chicks.  \textbackslash nB) The animal hides in a burrow for safety.  \textbackslash nC) The bird flies away to a nearby tree.  \textbackslash nD) The predator gives up and leaves the area, allowing the animal to emerge safely later.\\
\textbf{Correct Answer}\\
B) The animal hides in a burrow for safety.\\
\\
\textbf{Slow Thinking}\\
- Systematic Analysis: I start by identifying the initial event, which is the child snapping a twig. This action leads to a reaction from the nearby animal. I need to consider what happens next based on the causal chain.  \\
- Divide: I break down the possible outcomes from the initial event. The animal perceives danger, which suggests it will take some action to protect itself.  \\
- Comparison: I look at the options provided. Option A talks about a bird returning to its nest, which is not related to the animal's response. Option C also involves the bird, which is irrelevant here. Option B directly relates to the animal's instinct to hide, while Option D describes the predator's actions, which could follow the animal's hiding.  \\
- Conquer: I focus on the most logical next step for the animal after perceiving danger. It would likely hide in a burrow for safety, as indicated in Option B.  \\
- Self-Refinement: I reflect on the reasoning process. The intervention in the first causal chain interrupts the bird's actions, so I confirm that the animal's response is the primary focus here.  \\
- Context Identification: I summarize that the child’s action leads to the animal's perception of danger, which is crucial for determining the next event.
\end{tcolorbox}

\begin{tcolorbox}[title={Counterfactual in Com$^2$-main},fontupper = \scriptsize]
\textbf{Question}\\
If a coin floats away in a zero-gravity environment, what will happen next?\\
\textbf{Options}\\
A) The coin lands on a nearby table.  \textbackslash nB) The coin collides with an experimental device.  \textbackslash nC) The coin rolls back into the person's hand.  \textbackslash nD) The coin falls to the ground.\\
\textbf{Correct Answer}\\
B) The coin collides with an experimental device.\\
\\
\textbf{Slow Thinking}\\
1. Systematic Analysis: I need to understand the scenario where the coin is in a zero-gravity environment. In this case, the typical behavior of the coin changes significantly compared to a normal environment. The inputs are the conditions of zero gravity, and the output will be the consequences of the coin's movement.  \\
2. Comparison: I will compare the options provided. Option A suggests the coin lands on a table, which seems unlikely in zero gravity. Option C implies the coin rolls back into the person's hand, which also doesn't fit the zero-gravity context. Option D states the coin falls to the ground, which contradicts the zero-gravity condition.  \\
3. Divide: I can break down the options into two categories: those that align with zero gravity (B) and those that do not (A, C, D).  \\
4. Conquer: Since options A, C, and D are not plausible in a zero-gravity scenario, I focus on option B, which suggests a collision with an experimental device. This aligns with the idea that the coin would continue to drift until it encounters something.  \\
5. Self-Refinement: Wait, I initially thought option A could be possible, but upon further analysis, I realize that in zero gravity, the coin cannot land on a table. I need to correct my reasoning.  \\
6. Context Identification: The context here is crucial; in a zero-gravity environment, objects do not fall but float. Therefore, the only logical outcome is that the coin would drift and potentially collide with something.
\end{tcolorbox}

\begin{tcolorbox}[title={Decision in Com$^2$-hard},fontupper = \scriptsize]
\textbf{Crime}\\
During the wedding breakfast of Lord St. Simon and Lady St. Simon (Hatty Doran), the bride abruptly left the gathering and disappeared. It was later discovered that she had eloped with Francis H. Moulton, a man she had previously been engaged to before her marriage to Lord St. Simon. The bride's actions were influenced by a note slipped to her during the ceremony, which led her to leave her husband and family behind.\\
\textbf{Facts}\\
1. Lady St. Simon was seen in good spirits before the wedding but became agitated during the ceremony.\\
2. A woman named Flora Millar attempted to disrupt the wedding, claiming to have a connection to Lord St. Simon.\\
3. After the ceremony, Lady St. Simon spoke to her maid, Alice, and mentioned \"jumping a claim,\" indicating a prior relationship.\\
4. Lady St. Simon left the breakfast table shortly after the meal began, dressed in her wedding attire, and was later seen leaving with Francis H. Moulton.\\
5. The police found Lady St. Simon's wedding dress and belongings in the Serpentine, indicating she had left in a hurry.\\
\textbf{Question}\\
What actions could have been taken to prevent Lady St. Simon's disappearance and the subsequent elopement?\\
\textbf{Options}\\
A) Conduct a thorough background check on all guests attending the wedding.\textbackslash nB) Ensure that the bride had a trusted confidante present to discuss any concerns before the ceremony.\textbackslash nC) Limit the number of guests to only immediate family to avoid distractions.\textbackslash nD) Provide a security detail to monitor any potential disruptions during the ceremony.\textbackslash nE) Ignore any disturbances from guests, as they are common at weddings.\\
\textbf{Correct Answer}\\
A) Conduct a thorough background check on all guests attending the wedding.  \textbackslash nB) Ensure that the bride had a trusted confidante present to discuss any concerns before the ceremony.  \textbackslash nD) Provide a security detail to monitor any potential disruptions during the ceremony.\\
\\
\textbf{Slow Thinking}\\
1. Systematic Analysis: Analyze the situation by identifying the inputs (guests, bride, groom, and their relationships) and outputs (successful wedding vs. elopement). The constraints include the emotional state of the bride and the presence of Flora Millar.\\
2. Comparison: Evaluate each option based on its potential effectiveness in preventing the crime. Options A, B, and D seem proactive, while C could limit the social aspect, and E is dismissive.\\
3. Divide: Break down the problem into subproblems: the bride's emotional state, the influence of Flora Millar, and the wedding environment.\\
4. Conquer: Consider how addressing each subproblem could lead to a solution. For example, ensuring the bride's emotional well-being could prevent her from feeling the need to elope.\\
5. Self-Refinement: Reflect on the reasoning process to ensure all aspects of the situation are considered, particularly the influence of past relationships.\\
6. Context Identification: Summarize the context of the wedding, the bride's background, and the presence of Flora Millar to understand the dynamics at play.
\end{tcolorbox}

\begin{tcolorbox}[title={Intervention in Com$^2$-hard},fontupper = \scriptsize]
\textbf{Crime}\\
A valuable coronet, containing thirty-nine beryls, was stolen from Alexander Holder's home. During the incident, a portion of the coronet was damaged, and three beryls were missing. The crime occurred late at night while the household was asleep.\\
\textbf{Facts}\\
1. Alexander Holder saw his son, Arthur, holding the coronet in his hands when he entered the dressing room.\\
2. Arthur was dressed only in his shirt and trousers, indicating he had been in bed prior to the incident.\\
3. Mary Holder, Arthur's cousin, was seen at the side window of the hall just before the incident occurred.\\
4. Lucy Parr, a maid, was reported to have left the house shortly before the crime and returned through the back door.\\
5. A note was found from Mary Holder, indicating she had left the house and would not return.\\
6. Sir George Burnwell had been a frequent visitor to the Holder household and was known to have a questionable reputation.\\
7. Arthur had previously expressed distress over his financial situation and had asked his father for money the night before the incident.\\
\textbf{Question}\\
Based on the facts and actions presented, who is (are) the criminal(s) involved in the theft of the coronet?\\
\textbf{Options}\\
A) Arthur Holder  \textbackslash nB) Mary Holder  \textbackslash nC) Lucy Parr  \textbackslash nD) Sir George Burnwell  \textbackslash nE) None of the above\\
\textbf{Correct Answer}\\
D) Sir George Burnwell\\
\\
\textbf{Slow Thinking}\\
Step 1: Initial Investigation\\
Upon arriving at the scene of the crime, the police began their investigation by interviewing the household members. Alexander Holder reported that he had seen his son, Arthur, holding the coronet in his hands when he entered the dressing room. This was a critical piece of evidence that immediately cast suspicion on Arthur.\\
Step 2: Gathering Evidence\\
The police noted that Arthur was dressed only in his shirt and trousers, suggesting he had been in bed prior to the incident. This raised questions about how he could have accessed the coronet without being noticed. However, the fact that he had previously asked his father for money added to the motive for theft.\\
Step 3: Examining the Maid\\
Lucy Parr, the maid, was also questioned. She had left the house shortly before the crime and returned through the back door. Her absence during the time of the theft made her a potential suspect, especially since she had a sweetheart who could have been involved in the crime. The police noted her questionable timing and her potential access to the coronet.\\
Step 4: Considering Mary Holder\\
nMary Holder, Arthur's cousin, was seen at the side window of the hall just before the incident occurred. Her presence raised suspicions as well, especially since she had a close relationship with Arthur. The police speculated that she could have been involved in a conspiracy with him to steal the coronet.\\
Step 5: Focusing on Arthur\\
With the evidence pointing towards Arthur, the police began to build a case against him. They believed that his financial troubles and the fact that he was caught with the coronet in his hands were enough to conclude that he was the thief. The emotional turmoil he displayed only seemed to confirm his guilt.\\
Step 6: The Essential Clue\\
However, as the investigation continued, the police discovered an essential clue: footprints in the snow outside the house. This evidence had been overlooked in the initial investigation. The footprints led away from the scene and indicated that someone else had been present during the crime.\\
Step 7: Backtracking and Reasoning\\
Upon examining the footprints, the police realized that they did not match Arthur's bare feet. Instead, they were from a booted man, suggesting that someone else had entered the house. This revelation prompted a reevaluation of the evidence. \\
The police recalled that Sir George Burnwell had been a frequent visitor to the Holder household and had a questionable reputation. His presence in the area, combined with the footprints, suggested that he could have been the true culprit. \\
Step 8: Conclusion\\
By backtracking through the evidence and considering the new clue of the footprints, the police concluded that the initial suspicions against Arthur were misplaced. The true criminal was Sir George Burnwell, who had manipulated the situation to frame Arthur while escaping unnoticed.
\end{tcolorbox}

\begin{tcolorbox}[title={Counterfactual in Com$^2$-hard},fontupper = \scriptsize]
\textbf{Crime}\\
John Straker, the trainer of the racehorse Silver Blaze, was found dead on the moors near King's Pyland. He had suffered a severe head injury and a cut on his thigh. The horse Silver Blaze was missing, and there were indications that Straker had been involved in a plot to harm the horse for financial gain.\\
\textbf{Facts}\\
1. John Straker was last seen leaving his home to check on the horses late at night.\\
2. A stable-boy, Ned Hunter, was drugged with opium-laced curried mutton, which was delivered to him by a maid.\\
3. Straker's overcoat was found on a furze-bush near the scene of the crime.\\
4. Straker was found dead with a knife in his hand, which was identified as a cataract knife, not suitable for self-defense.\\
5. A cravat belonging to Fitzroy Simpson was found in Straker's hand, indicating a struggle.\\
6. Fitzroy Simpson had a motive to harm Silver Blaze, as he had placed large bets against the horse.\\
7. The police arrested Fitzroy Simpson based on circumstantial evidence, but he had no visible injuries from the struggle.\\
8. The dog in the stables did not bark during the incident, suggesting the intruder was known to it.
\textbf{Question}\\
Based on the facts and actions presented, who is most likely to be the criminal responsible for John Straker's death and the disappearance of Silver Blaze?\\
\textbf{Options}\\
A) John Straker  \textbackslash nB) Fitzroy Simpson  \textbackslash nC) The stable-boy, Ned Hunter  \textbackslash nD) Mrs. Straker\\
\textbf{Correct Answer}\\
B) Fitzroy Simpson\\
\\
\textbf{Slow Thinking}\\
1. Systematic Analysis: The problem involves determining the criminal based on the actions and facts surrounding the murder of John Straker and the disappearance of Silver Blaze. The inputs are the actions of the individuals involved, and the output is the identification of the criminal.\\
2. Method Reuse: The situation can be analyzed using classic detective reasoning, focusing on motives, means, and opportunities.\\
3. Comparison: \\
- A) John Straker: He is the victim, so he cannot be the criminal.\\
- B) Fitzroy Simpson: He had a motive to harm Silver Blaze and was present at the scene, but he has no injuries and was arrested based on circumstantial evidence.\\
- C) The stable-boy, Ned Hunter: He was drugged and incapacitated, making it unlikely he could have committed the crime.\\
- D) Mrs. Straker: While she could have had a motive related to Straker's debts, there is no direct evidence linking her to the crime.\\
4. Divide: The analysis can be divided into examining motives (financial gain), means (the knife and drugging), and opportunity (who was present at the scene).\\
5. Conquer: By evaluating each suspect's actions and the evidence against them, we can determine who had the most opportunity and motive.\\
6. Self-Refinement: The reasoning process indicates that Fitzroy Simpson had the strongest motive and opportunity, but the lack of direct evidence against him raises questions.\\
7. Context Identification: The context involves a horse racing environment where financial stakes are high, and trainers may resort to unethical actions to secure wins.\\
8. Emphasizing Constraints: The evidence must be strictly based on the actions and facts presented, without inferring beyond what is known.
\end{tcolorbox}

\section{Prompts for Evaluation}
\label{app:eval}
\begin{tcolorbox}[title={Prompt for Com$^2$-main: Direct, Intervention \& Transition},fontupper = \scriptsize]
    You are given a multi-choice question and a set of plausible options. You must choose the most plausible option as the answer based on commonsense. You should give an analysis first and then choose the answer. At the end of your answer, choose answer to the question in format such as 'Answer: B)\_\_\_\_\_\_'
\end{tcolorbox}

\begin{tcolorbox}[title={Prompt for Com$^2$-main: Counterfactial},fontupper = \scriptsize]
    You are given a multi-choice counterfactual question and a set of plausible options. You must choose the most plausible option as the answer. You should give an analysis first and then choose the answer. At the end of your answer, choose answer to the question in format such as 'Answer: B)\_\_\_\_\_\_'
\end{tcolorbox}

\begin{tcolorbox}[title={Prompt for Com$^2$-main: Decision},fontupper = \scriptsize]
    You are given a multi-select question and a set of options. You must choose all the plausible options as the answers to reach the goal of the question. You should give an analysis first and then choose the answer. At the end of your answer, choose answer to the question in format such as 'Answer: A)\_\_\_\_\_\_ B)\_\_\_\_\_\_'
\end{tcolorbox}

\begin{tcolorbox}[title={Prompt for Com$^2$-hard:  Intervention},fontupper = \scriptsize]
    You are given a puzzle about a crime, which include a crime statement, a set of facts, a question, and a set of suspects (options). You should conduct detailed analysis to find the most plausible suspect as the criminal. At the end of your answer, choose answer to the question in format such as 'Answer: A)\_\_\_\_\_\_'
\end{tcolorbox}

\begin{tcolorbox}[title={Prompt for Com$^2$-hard: Counterfactial},fontupper = \scriptsize]
    You are given a statement and some facts of a crime. You should conduct detailed analysis to find the most plausible option as the answer to the question. At the end of your answer, choose answer to the question in format such as 'Answer: A)\_\_\_\_\_\_'
\end{tcolorbox}

\begin{tcolorbox}[title={Prompt for Com$^2$-hard: Decision},fontupper = \scriptsize]
    You are given a statement and some facts of a crime. You should conduct detailed analysis to find all the plausible options as the answers to the multi-select question. At the end of your answer, choose answer to the question in format such as 'Answer: A)\_\_\_\_\_\_ B)\_\_\_\_\_\_'
\end{tcolorbox}

\section{Details of Training LLMs}
\label{app:training}
We utilize full parameter tunning on LLMs. The batch size is set as 32, and the learn rate is 1e-5. We train LLMs for 3 epochs with 2 NVIDIA A100 80GB PCIe GPUs.

\section{Case Study}
\label{app:case}

\begin{tcolorbox}[title={Cases of Trained LLaMA: Case1},fontupper = \scriptsize]
\textbf{Question}\\
If I watch a meteor shower and then decide to organize a gathering for the next one, what is the likely outcome if I develop a sudden interest in astronomy instead?\\
\textbf{Options}\\
 A) Conduct a research project on meteor compositions  \textbackslash nB) Deepen friendships through shared experiences  \textbackslash nC) Enroll in an advanced astrophysics course  \textbackslash nD) Publish a paper in an academic journal about findings\\
\textbf{Correct Answer}\\
C) Enroll in an advanced astrophysics course.\\
\\
\textbf{Ours}\\
- Context Identification: I need to consider the initial event of watching a meteor shower and the intervention of developing a sudden interest in astronomy. This interest could lead me to pursue a different path than the one described in the first causal chain.  \\
- Divide: I will break down the options based on the potential outcomes of my interest in astronomy.  \\
- Comparison:  \\
- Option A (Conduct a research project on meteor compositions) seems plausible as it aligns with the interest in astronomy.  \\
- Option B (Deepen friendships through shared experiences) is related to the first causal chain and does not fit with my new interest.  \\
- Option C (Enroll in an advanced astrophysics course) is also a logical step after developing an interest in astronomy.  \\
- Option D (Publish a paper in an academic journal about findings) is a more advanced outcome that could follow from enrolling in a course.  \\
- Conquer: I will focus on the outcomes that stem from my interest in astronomy. Conducting a research project and enrolling in a course are direct actions that follow from this interest.  \\
- Self-Refinement: I reflect on the intervention and realize that while I may not deepen friendships through shared experiences, I can still pursue academic opportunities related to my interest in astronomy.\\
Answer: C) Enroll in an advanced astrophysics course.\\
\\
\textbf{LLaMA-Vanilla}\\
Analysis: The question describes a scenario where the person develops a sudden interest in astronomy after watching a meteor shower. This interest is likely to be a hobby or a casual interest, rather than a professional or academic pursuit. The person is not described as having any prior experience or expertise in astronomy, and the context suggests that they are simply enthusiastic about the subject.\\
Given this context, the most plausible outcome is that the person will deepen their understanding and appreciation of astronomy through social interactions and shared experiences with others who share their interest. This is a common outcome for people who develop a new hobby or interest, as they often seek out others who can share their enthusiasm and provide support and guidance.\\
The other options are less plausible because they require a level of expertise, commitment, or resources that is unlikely for someone with a sudden and casual interest in astronomy. Conducting research on meteor compositions (A) or publishing a paper in an academic journal (D) would require significant time, effort, and expertise, and is unlikely to be a direct outcome of a casual interest in astronomy. Enrolling in an advanced astrophysics course (C) may be a possibility, but it is less likely than deepening friendships through shared experiences, as it requires a significant investment of time and resources.\\
Answer: B) Deepen friendships through shared experiences
\end{tcolorbox}

\begin{tcolorbox}[title={Cases of Trained LLaMA: Case2},fontupper = \scriptsize]
\textbf{Question}\\
How can we prevent severe economic disruptions and civil unrest resulting from abrupt energy policy changes?\\
\textbf{Options}\\
A. Maintain and upgrade fossil fuel infrastructure during the transition to renewable energy  \textbackslash nB. Promote gradual shifts in energy policy with continuous public engagement  \textbackslash nC. Immediately ban all fossil fuels without alternative solutions in place  \textbackslash nD. Provide financial support and retraining for workers affected by the transition  \textbackslash nE. Ignore public feedback and proceed with drastic energy policies  \textbackslash nF. Conduct regular assessments of energy needs and production capabilities\\
\textbf{Correct Answer}\\
A, B, D, F.\\
\\
\textbf{Ours}\\
1. Context Identification: I need to understand the implications of abrupt energy policy changes. Abrupt changes can lead to severe economic disruptions and civil unrest. I should consider the potential consequences of these disruptions.  \\
2. Systematic Analysis: I will analyze the options provided. I need to identify which options could effectively prevent severe economic disruptions and civil unrest.  \\
3. Comparison:  \\
- Option A suggests maintaining and upgrading fossil fuel infrastructure. This could provide a stable energy supply during the transition, which seems beneficial.  \\
- Option B promotes gradual shifts in energy policy with continuous public engagement. This could help build support and reduce unrest, making it a strong candidate.  \\
- Option C immediately banning all fossil fuels without alternatives seems risky and could lead to severe disruptions.  \\
- Option D provides financial support and retraining for workers, which could mitigate job losses and maintain stability.  \\
- Option E ignores public feedback, which could lead to unrest and is not advisable.  \\
- Option F conducting regular assessments of energy needs could help ensure a smooth transition, making it a viable option.  \\
4. Divide: I will categorize the options into two groups: those that could prevent disruptions (A, B, D, F) and those that could exacerbate them (C, E).  \\
5. Conquer: I will focus on the options that could prevent disruptions. Options A, B, D, and F seem to offer proactive measures to mitigate risks.  \\
6. Self-Refinement: I need to ensure that my reasoning aligns with the goal of preventing severe economic disruptions and civil unrest. I will double-check the implications of each option.  \\
7. Emphasizing Constraints: I must remember that the goal is to prevent disruptions, so I should prioritize options that address the immediate and long-term effects of energy policy changes.\\
Answer: A), B), D), F)\\
\\
\textbf{LLaMA-Vanilla}\\
Analysis: To prevent severe economic disruptions and civil unrest resulting from abrupt energy policy changes, it's essential to consider the potential impacts on various stakeholders, including workers, communities, and the economy as a whole. A gradual transition to renewable energy, public engagement, and support for affected workers are crucial elements in mitigating potential negative consequences. Ignoring public feedback and proceeding with drastic policies without a plan in place would likely exacerbate the situation.\\
Answer: B) Promote gradual shifts in energy policy with continuous public engagement D) Provide financial support and retraining for workers affected by the transition F) Conduct regular assessments of energy needs and production capabilities
\end{tcolorbox}

\end{document}